\documentclass{article}

\usepackage{arxiv}

\usepackage[utf8]{inputenc} 
\usepackage[T1]{fontenc}    
\usepackage{hyperref}       
\usepackage{url}            
\usepackage{booktabs}       
\usepackage{amsfonts}       
\usepackage{amsmath}
\usepackage{nicefrac}       
\usepackage{microtype}      
\usepackage{lipsum}		
\usepackage{graphicx}
\usepackage{caption}
\usepackage{subcaption}
\usepackage{natbib}
\usepackage{xcolor}
\usepackage{doi}
\usepackage{algorithm}
\usepackage{algpseudocode}

\newcommand{\R}{\mathbb{R}}
\newcommand{\calD}{\mathcal{D}}

\newcommand{\qref}[1]{Eq.~(\ref{eqn:#1})}
\newcommand{\sref}[1]{Sec.~\ref{#1}}
\newcommand{\fref}[1]{Fig.~(\ref{fig:#1})}
\newcommand{\aref}[1]{Alg.~(\ref{alg:#1})}

\usepackage{amsthm}

\title{Physics-informed Information Field Theory for Modeling Physical Systems With Uncertainty Quantification}

\author{{\includegraphics[scale=0.06]{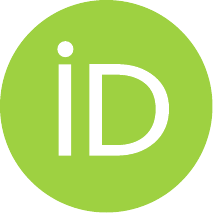}\hspace{1mm}Alex Alberts} \\
	School of Mechanical Engineering\\
	Purdue University\\
	West Lafayette, IN \\
	\texttt{albert31@purdue.edu} \\
	\And
    {\includegraphics[scale=0.06]{orcid.pdf}\hspace{1mm}Ilias Bilionis}\footnote{Corresponding author} \\
	School of Mechanical Engineering\\
	Purdue University\\
	West Lafayette, IN \\
	\texttt{ibilion@purdue.edu} \\
}

\renewcommand{\shorttitle}{Physics-informed Information Field Theory}

\hypersetup{
  pdftitle={Physics-informed_Information_Field_Theory},
  pdfauthor={Alex Alberts, Ilias Bilionis}
}

\begin{document}
\maketitle

\begin{abstract}
Data-driven approaches coupled with physical knowledge are powerful techniques to model engineering systems. The goal of such models is to efficiently solve for the underlying physical field through combining measurements with known physical laws. As many physical systems contain unknown elements, such as missing parameters, noisy measurements, or incomplete physical laws, this is widely approached as an uncertainty quantification problem. The common techniques to handle all of these variables typically depend on the specific numerical scheme used to approximate the posterior, and it is desirable to have a method which is independent of any such discretization. Information field theory (IFT) provides the tools necessary to perform statistics over fields that are not necessarily Gaussian. The objective of this paper is to extend IFT to physics-informed information field theory (PIFT) by encoding the functional priors with information about the physical laws which describe the field. The posteriors derived from this PIFT remain independent of any numerical scheme, and can capture multiple modes which allows for the solution of problems which are not well-posed. We demonstrate our approach through an analytical example involving the Klein-Gordon equation. We then develop a variant of stochastic gradient Langevin dynamics to draw samples from the field posterior and from the posterior of any model parameters. We apply our method to several numerical examples with various degrees of model-form error and to inverse problems involving non-linear differential equations. As an addendum, the method is equipped with a metric which allows the posterior to automatically quantify model-form uncertainty. Because of this, our numerical experiments show that the method remains robust to even an incorrect representation of the physics given sufficient data. We numerically demonstrate that the method correctly identifies when the physics cannot be trusted, in which case it automatically treats learning the field as a regression problem.
\end{abstract}


\section{Introduction}
\label{sec:intro}

We introduce \emph{physics-informed information field theory} (PIFT), an extension of information field theory (IFT) \cite{ensslin2009information,ensslin2013information}.
IFT treats learning a field of interest in a fully Bayesian way which makes use of concepts from statistical field theory.
IFT is a general method for solving forward or inverse problems involving physical systems that incorporates multiple sources of uncertainty along with any known physics in a unified Bayesian way.
In \cite{ensslin2022information}, a concise introduction to IFT is presented along with the connection between IFT and artificial intelligence.
Typically in IFT, the functional priors are Gaussian random fields which encode known regularity constraints on the field.
To extend IFT to PIFT, we derive physics-informed functional priors which are encoded with knowledge of the physical laws that the field obeys.
Furthermore, PIFT is equipped with a device which allows for detection of an incorrect representation of the physics.
Because of this, PIFT is self-correcting, as it can automatically balance the contributions from the physics and any available measurements to the form of the posterior.

Previous approaches for modeling systems from physical knowledge and measurements with uncertainty typically need to combine multiple methods together.
For example, combine a numerical solver for the field, such as the physics-informed neural networks (PINNs) \cite{raissi2017physics}, with an additional method for uncertainty quantification, e.g. Bayesian PINNs \cite{yang2021b}.
However, in the case of PIFT we derive the posterior over the entire field of interest, and since PIFT is derived from classic IFT, the posterior is independent of any numerical schemes.
The PIFT posterior contains an infinite number of degrees of freedom, and it is defined over the entire field of interest in a continuous manner, rather than as an approximation of the field.
Treating the field in this probabilistic way allows PIFT to handle a wide class of problems.
PIFT can be used to approach forward problems, inverse problems where the sources of uncertainty are combined seamlessly, and can even find the correct solutions to problems which are not well-posed by capturing multiple modes each representing a possible solution to the problem.
In a sense, PIFT should serve as a starting point for deriving approximate methods for physics-informed modeling approaches.

In \sref{sec:pinns}, we provide a brief summary of the state-of-the-art approaches and other methods which are related to PIFT.
In \sref{sec:ift}, we familiarize the readers with classic IFT and demonstrate how probability measures over function spaces can be defined for use in Bayesian inference.
Then, in \sref{sec:methodology}, we introduce PIFT as a way to incorporate knowledge of physical laws into IFT in the form of physics-informed functional priors for forward problems in \sref{sec:methods} and inverse problems in \sref{inverse}. A case where an analytic representation of the posterior is available is shown in \sref{sec:analytic}.
In \sref{sec:numericalposts}, a stochastic gradient Langevin dynamics sampling scheme for PIFT forward problems is developed, which can be slightly modified for inverse problems.
The relationship between PIFT and PINNs is discussed in \sref{sec:relation_to_pinns}, where we observe that classic PINNs and Bayesian PINNs are both special cases of the PIFT posterior.
Finally, in \sref{sec:examples} we provide numerical experiments which showcase various characteristics of the method.

\section{Review of state-of-the-art approaches}
\label{sec:pinns}
Physics-informed modeling approaches have been shown to be powerful methods for modeling physical systems.
These methods combine data-driven approaches with physics-based modeling techniques to improve the efficiency of the learning process.
By training the model of the system with some physical knowledge, accurate results can be achieved with limited data, or even in some cases without data \cite{stiasny2021learning}.
As most physical systems are known to obey specific differential equations, typically physics-informed models are built by representing the field with a parameterization which is trained to obey both the measurements and the differential equations simultaneously.

One of the most ubiquitous choices for surrogate functions in physics-informed modeling are neural networks.
The idea to use neural networks to solve differential equations was first proposed as far back as the 1990s in \cite{psichogios1992hybrid,meade1994numerical}, and most notably due to the similarities to modern methods~\cite{lagaris1998artificial}, where the authors train a neural network to learn the solutions of partial differential equations.
With deep neural networks (DNNs) seeing a major rise in popularity, this idea was revisited and formalized in \cite{raissi2017physics} using PINNs to learn the solution of a partial differential equation (PDE) characterizing physical systems as well as automating the discovery of the underlying PDE from noisy measurements of the system.
Generally, the idea behind PINNs is to parameterize state variables in the system as deep neural networks and forcing the network to respect any applicable physical laws using additional regularizers in the form of PDE residuals. 
These additional regularizers have the effect of adding physical constraints on the prior over the space of DNNs leading to efficient learning under limited data. 

Due to the generic nature of PINNs, they have seen success in a wide number of applications. 
Some of these include fluid mechanics \cite{raissi2018hidden}, biomedical applications \cite{10.3389/fphy.2020.00042, KISSAS2020112623}, material identification \cite{zhang2020physicsinformed}, heat transfer \cite{cai2021physics}, magnetostatic problems \cite{beltran2022physics}, and many other applications. 
Much of the research involving PINNs has a focus on discovering underlying physics using neural networks. 
For example, the governing equations for a system might not be known exactly, or some parameters in the problem may be unknown.
PINNs can be used to estimate or discover these relationships.
\cite{tartakovsky2018learning} leverage the power of PINNs to learn constitutive laws in nonlinear diffusion PDEs. 
A recent application on automatic data-driven discovery of physical laws can be found in \cite{PhysRevLett.124.010508}, where the task of discovering physical laws (such as governing equations) is formalized as a representation learning problem. 
In \cite{lu2019deeponet}, the authors developed an approach for learning nonlinear operators for identifying differential equations called \textit{DeepONet}.

Because most real-world physical systems have stochastic elements in them, there is much interest in combining physics-informed modeling with techniques from uncertainty quantification.
The classic approaches seek to combine an existing numerical solver, such as a finite element method, with a method for uncertainty quantification \cite{marelli2014uqlab}.
This idea has been extended to PINNs.
In \cite{karumuri2019simulator}, elliptic PDEs whose coefficients are random fields are studied.
A PINN is built to do uncertainty quantification of the output.
The energy functional of the PDE is used in place of the residual due to the better behavior and faster convergence seen.
A data-free approach for physics-informed surrogate modeling for stochastic PDEs with high-dimensional varying coefficients is presented in \cite{zhu2019physics}.
A probabilistic surrogate model using PINNs combined with Bayesian inference is built. 
This model is approximated through variational inference, choosing a Boltzmann-Gibbs distribution as the reference density \cite{graves2011practical, blundell2015weight}.
In addition to these, there are a number of non-Bayesian approaches that have been developed that can quantify uncertainty for PINNs such as dropout, \cite{gal2016dropout}, physics-informed generative adversarial networks (PI-GANs) \cite{yang2020physics}, and polynomial chaos expansions combined with dropout \cite{zhang2019quantifying}.

A recent probabilistic approach is the Bayesian physics-informed neural networks (B-PINNs) proposed in \cite{yang2021b}, which are used to solve PDEs with noisy data for forward and inverse problems. 
B-PINNs consist of a Bayesian neural network, which is a neural network that puts a prior over the network parameters \cite{mackay1992practical,neal2012bayesian}.
They add a standard measurement likelihood with a fictitious likelihood that assumes measurement of zero PDE residuals.
Then, Hamiltonian Monte Carlo \cite{neal2011mcmc}, \cite{neal2012bayesian} or variational inference \cite{blei2017variational} are used to estimate the posterior distribution.
Probabilistic approaches for solving PDEs with periodic boundary conditions in a similar way to B-PINNs are presented in \cite{bilionis2016} and in \cite{frank2020probabilistic}.
Like the B-PINNs approach, the physics are encoded via discrete measurements of the PDE residual through the likelihood.
Instead of discretizing the field with a B-PINN, IFT is used to perform inference over the field directly.
In \cite{chen2021solving}, Gaussian processes are employed to learn the solutions of PDEs.
This approach uses the maximum a posterior estimator of a Gaussian process which is conditioned on the solution of the PDE at a finite grid of collocation points to solve the problem.
They argue that the Gaussian process approach has advantages over the PINNs approach because the theory of Gaussian processes is well-understood, unlike the case for PINNs.

Finally, we discuss some approaches that employ functional priors when doing inference, a concept which is related to information field theory.
The first such approach can be found in~\cite{meng2021learning}.
The problem of modeling a physical system is treated through Bayesian inference, which allows for uncertainty quantification over the field of interest.
First, the PI-GANs are used to learn a functional prior from previous data with physics, or from a prescribed functional distribution such as a Gaussian process.
The physics are encoded either through PINNs, or by learning the corresponding operator with DeepONet.
Once this functional prior is learned, Hamiltonian Monte Carlo is used to sample from the posterior in the latent space of PI-GANs.
In \cite{cotter2010approximation} a Bayesian field theory approach is used to approximate the solutions of inverse problems.
Similarly to IFT, an infinite-dimensional functional prior is employed when doing inference over the space of solutions.
However, the priors are defined over the space of inputs, such as a spatially-dependent thermal conductivity when solving the heat equation, rather than over the space of solutions to the PDE itself as done in IFT.
Furthermore this approach considers only Gaussian random field priors.
Finally, the work of \cite{dashti2013map} is an extension of \cite{cotter2010approximation} which theoretically studies the existence and well-defined nature of MAP estimators for the posteriors given in their work.
 
\section{Review of information field theory}
\label{sec:ift}
To develop a probabilistic approach to physics-based modeling we will lean heavily on the machinery of IFT~\cite{ensslin2013information}. 
In much of the literature involving this intersection of probability with physics-informed modeling, many probability measures are defined without consideration as to whether or not the measure itself is \emph{well-defined}.
While we have observed these algorithms to perform well in practice, this crucial detail could prevent a method from generalizing if many of the restrictions are relaxed, such as Gaussian probabilities.
Furthermore, in the usual approaches the posteriors derived are dependent on the specific choice of parameterization of the field.
IFT provides a framework in which we can define probability measures over the field itself in a theoretically sound way, which removes any dependence on parameterizations.
Specifically, IFT provides a natural way to define probability measures directly over the field of interest in a way which is similar to \cite{meng2021learning}.
However, instead of trying to learn a functional prior which describes the field, we can define a specific form of the functional prior from IFT that honors the physics.
In addition to this, IFT grants analytical representations of the statistics of the field, or even an analytical form of the posterior, in certain cases.

In a simple sense, IFT is the application of information theory \cite{kullback1997information} to fields, where in this case the chosen field refers to the physical system of interest.
IFT was first presented in \cite{ensslin2009information} as a Bayesian statistical field theory for non-linear image reconstruction and signal analysis problems, and provides a methodology for constructing signal recovery algorithms for non-linear and non-Gaussian signal inference problems, where the posterior contains an infinite number of degrees of freedom. 
Similar to a PINN, IFT allows information about constraints that the field is known to obey, such as physical laws, statistical symmetries \cite{frewer2016note}, or smoothness properties, to be encoded as prior information.
Then this prior information about the field is combined with data through Bayesian inference. 
We remark that IFT is simply an application of statistical field theory, with which many physicists are already familiar~\cite{parisi1988statistical}.

In general, IFT is concerned with learning a field of interest $\phi$ defined over a physical domain $\Omega$.
We denote by $x$ points in $\Omega$ and by $\phi(x)$ the value of the field $\phi$ at physical point $x$.
Typically, the function $\phi$ maps $\Omega$ to the real line $\mathbb{R}$ and it lives in a Hilbert space $\Phi$.
We define a probability measure over $\Phi$, i.e., a functional prior.
Formally, we denote the density of this functional prior by $p(\phi)$.
With this prior, we can define the field statistics, e.g., the expectation of a functional $f$ from $\Phi$ to $\mathbb{R}$ is:
\begin{equation}
    \mathbb{E}[f[\phi]] = \int_{\Phi} \calD\phi\: f[\phi]p(\phi). \label{eqn:expect_prior}
\end{equation}

The integral in \qref{expect_prior} is a path integral (sometimes called a functional integral) \cite{cartier2006functional}.
Path integrals are commonly encountered in SFT and in quantum field theory \cite{parisi1988statistical, lancaster2014quantum}.
The domain of integration is not a set of points over the real or complex domain, but a function space.
Intuitively, the path integral in \qref{expect_prior} can be interpreted as follows: sum the value of the functional $f[\phi]$ for each function $\phi$ in the function space $\Phi$.
Typically, functional integrals must be computed numerically, although in some simple cases analytic solutions are available.

To demonstrate how IFT works in practice, we cover the simplest example case, which is referred to as a \emph{free theory}.
To this end, define the inner product, $\psi^\dagger\phi$, of two functions $\psi$ and $\phi$ in the function space $\Phi$ as:
$$
\psi^\dagger\phi \equiv \int_{\Omega}dx\:\phi(x)\psi(x).
$$
Now consider a bounded linear operator $S$ from $\Phi$ to itself.
Under the Schwartz kernel theorem \cite{treves2016topological}, the operator $S$ can be represented as an integral operator:
$$
(S\phi)(x) = \int dx'\: S(x, x')\phi(x').
$$
The inverse operator $S^{-1}$ satisfies:
$$
SS^{-1}\phi = \phi,
$$
for each $\phi$ in $\Phi$.
This equation is equivalent to:
$$
\int dx' dx''\: S(x, x')S^{-1}(x', x'')\phi(x'') = \phi(x).
$$
As we will show in \sref{sec:analytic}, under special assumptions for $S(x,x')$, $S^{-1}$ is a common Green's function.

We say that the operator $S$ is positive-definite if
$
\phi^\dagger S \phi > 0,
$
for all non-zero $\phi$ in $\Phi$.
It can be shown that $S$ is positive-definite if and only if its inverse $S^{-1}$ is positive-definite \cite{keener2018principles}.
Suppose $S$ is a positive-definite operator.
We use it to define a zero mean Gaussian probability measure over the function space $\Phi$.
The technical definition is involved \cite{albeverio1976mathematical}, but formally we can define the density of that probability measure by:
\begin{equation}
    \label{eqn:field_prior}
    p(\phi) \propto \exp\left\{-\frac{1}{2}\phi^\dagger S\phi\right\}.
\end{equation}

In what follows, we derive the first statistics of $\phi$.
We present the derivation used by physicists.
This approach is not mathematically rigorous, but it is more intuitive.
All concepts can be made mathematically rigorous using the theory of Gaussian random fields \cite{bardeen1986statistics}.

We start by introducing the \emph{partition function}:
\begin{equation}
\label{eqn:partition}
Z[q] = \int \calD\phi\: \exp\left\{-\frac{1}{2}\phi^\dagger S\phi + \phi^\dagger q\right\},
\end{equation}
for any function $q$ in $\Phi$.
Notice that $Z[q=0]$ is the normalization constant of \qref{field_prior}, and this normalization constant may be infinite.
We will see later on that it will cancel out when calculating the statistics of the field.

Now think of $q$ and $\phi$ as vectors and $S$ as a matrix.
Then \qref{partition} resembles a standard multivariate Gaussian integral.
From this observation, we can guess that the form of the partition function is:
\begin{equation}
Z[q] = C\left[\det S\right]^{-1/2}\exp\left\{\frac{1}{2} q^\dagger S^{-1}q\right\},
\label{eqn:partition_analytical}
\end{equation}
where $C$ is a constant and $\det S$ is the determinant of the operator (the product of all its eigenvalues).
Again, both $C$ and $\det S$ may be infinite, but they will cancel out.

From this definition, we can derive the expectation of the field $\phi$ at a point $x$, which is given by:
\begin{equation}
    \mathbb{E}[\phi(x)] = \int \calD\phi\: \frac{\exp\left\{-\frac{1}{2}\phi^\dagger S\phi\right\}}{Z[q=0]}\phi(x) = 0.
    \label{eqn:expect_of_field}
\end{equation}
Since this is a zero-mean field, the covariance is:
\begin{equation}
    \mathbb{E}[\phi(x)\phi(x')] = \int \calD\phi\: p(\phi) \phi(x)\phi(x') = S^{-1}(x,x').
    \label{eqn:cov_of_field}
\end{equation}
For derivations of \qref{expect_of_field} and \qref{cov_of_field} see Appendix~\ref{appendix:expectation_and_cov_prior}.
Higher order statistics can be found using Wick's theorem~\cite{lancaster2014quantum}.

We saw that the choice of $S$ corresponds to a choice of a covariance function $S^{-1}(x,x')$.
In IFT, one typically starts by choosing $S^{-1}(x,x')$ based on some prior beliefs about the regularity and the lengthscale of the field.
Under this interpretation, the free theory reverts to the theory of Gaussian random fields, otherwise known as the theory of Gaussian processes.
This justifies saying that $\phi$ follows a zero mean Gaussian process with covariance function $S^{-1}$ and writing
$$
\phi \sim \mathcal{N}(0, S^{-1}).
$$
Alternatively, and somewhat abusing the mathematical notation, we may write that the ``probability density'' of the field $\phi$ is:
$$
p(\phi) = \mathcal{N}(\phi | 0, S^{-1}).
$$

Next we construct the likelihood using $s$ measurements of the field.
These measurements are consolidated into a dataset $d = (d_1,\dots,d_s)$.
For simplicity, we assume that
$$
d = R\phi + n,
$$
where $R$ is an $s$-tuple of bounded linear functionals and $n = (n_1,\dots,n_s)$ is a noise process.
We call $R$ the \emph{measurement operator}.
Under these assumptions the well-known Riesz representation theorem applies, so each component functional $R_i$ of $R$ can be represented as an inner product  \cite{kreyszig1991introductory}:
$$
    R_i\phi = r_i^\dagger\phi = \int_{\Omega} dx\: r_i(x) \phi(x),
$$
for some element $r_i$ in the function space $\Phi$.
The noise $n$ is, typically, assumed to be a zero-mean, $s$-dimensional, Gaussian with covariance matrix $N$, i.e., $n\sim \mathcal{N}(0,N)$.
Concluding, the likelihood of the data is:
$$
    p(d|\phi) = \mathcal{N}(d|R\phi, N).
$$

The field posterior is constructed from Bayes's theorem:
\begin{equation}
    p(\phi|d) \propto p(d|\phi)p(\phi). \label{eqn:free_post}
\end{equation}
In order to understand the posterior, we can rewrite \qref{free_post} in the language of statistical field theory.
We write
$$
    p(\phi|d) \propto \exp\left\{-H[\phi|d]\right\}
$$
where 
$$
H[\phi|d] = \frac{1}{2}\left[\phi^\dagger S \phi + (R\phi - d)^T N^{-1} (R\phi - d)\right].
$$
is known as the \emph{information Hamiltonian}.
The resulting posterior is also Gaussian:
$$
    p(\phi|d) = \mathcal{N}(\phi|\tilde{m},\tilde{S}^{-1}).
    \label{free-posterior}
$$
The mean and the inverse of the covariance function can be found via completing the square.
The mean is:
\begin{equation}
\label{eqn:posterior_mean:v1}
\tilde{m} = \tilde{S}^{-1}R^{\dagger}N^{-1}d,
\end{equation}
and the inverse covariance operator is:
$$
\tilde{S} = S + R^\dagger N^{-1} R
$$
Here, $R^{\dagger}$ denotes the adjoint of $R$ defined by:
$$
\phi^\dagger R^\dagger = (R\phi)^T,
$$
or component wise:
$$
\phi^\dagger R_i^{\dagger} = \phi^\dagger r_i = \int dx\: \phi(x) r_i(x).
$$
This result is well-documented in Wiener filter theory \cite{ensslin2013information}.

The mean and covariance equations take a more familiar form if one uses the Sherman–Morrison–Woodbury formula \cite{deng2011generalization} (assuming all operators are bounded).
The formula yields an alternative expression for the covariance function:
\begin{equation}
\label{eqn:posterior_covariance}
\tilde{S}^{-1} = S^{-1} - S^{-1}R^\dagger(N + RS^{-1}R^\dagger)^{-1}RS^{-1},
\end{equation}
Notice that $K=N + RS^{-1}R^\dagger$ is an $s\times s$ positive-definite matrix with elements:
\begin{equation}
\label{eqn:covariance_matrix}
K_{ij} = N_{ij} + R_i^\dagger S^{-1} R_j =  N_{ij} + \int dx dx'\: r_i(x) S^{-1}(x,x') r_j(x').
\end{equation}
Using \qref{posterior_covariance} in \qref{posterior_mean:v1}, we get
\begin{equation}
\label{eqn:posterior_mean}
\tilde{m} = S^{-1}R^\dagger K^{-1}d.
\end{equation}

\qref{posterior_covariance} and \qref{posterior_mean} become the common formulas that one encounters in Gaussian process regression \cite{schulz2018tutorial} if the measurement operator picks the field at specific points.
To see this, take the measurement functionals to be Dirac delta functions centered at sensor locations $x_i$ in $\Omega$,
$$
r_i(x) = \delta(x - x_i),
$$
and observe that \qref{covariance_matrix} becomes the classical covariance matrix in Gaussian process regression:
$$
K_{ij} 
= N_{ij} + S^{-1}(x_i, x_j).
$$Similarly the terms $RS^{-1}$ and $S^{-1}R^\dagger$ become the familiar cross-covariance terms, e.g.,
$$
(R_iS^{-1})(x) = (r_i^\dagger S^{-1})(x) = \int dx'\: r_i(x')S^{-1}(x', x) = S^{-1}(x_i,x).
$$

While the definitions and examples here provided for IFT are for steady-state systems, IFT can be generalized to systems with dynamics through \emph{information field dynamics} \cite{ensslin2013information}.

\paragraph{On infinities.}
For the free theory, the partition function can be calculated analytically, but it is not necessarily finite.
The infinity arises from the fact that we are integrating over all functions, which may include functions with arbitrarily small scales.
To tame this infinity, physicists truncate the integration at the smallest scale of physical relevance via renormalization, see \cite{lancaster2014quantum}.
Renormalization is essential for developing perturbative approximations of non-free theories.
In practice, choosing a parameterization of $\phi$, e.g., expanding it in a basis or using a neural network, is equivalent to truncating the functional integration at some scale, which we make use of for this paper.

\section{Methodology}
\label{sec:methodology}
\subsection{Physics-informed information field theory: forward problems}
\label{sec:methods}

We begin by studying forward problems, where we only wish to derive a posterior over the field of interest.
As we will later see, inverse problems will require more care as the partition function will depend on the parameters trying to be inferred.
We first demonstrate how to define a physics-informed functional prior.
We make use of IFT to define a probability measure over a function space of interest that is embedded with the known physics of the field.
We then demonstrate how to incorporate observations of the field of interest in the form of the likelihood.
Finally, we derive the posterior over the solution field of the PDE from the physics-informed prior and likelihood.

\subsubsection{Physics-informed functional prior}
Bayesian inference over physical fields requires starting with a prior probability measure.
Our approach leverages IFT to construct such a probability measure with the property that fields which satisfy the physics are a priori more likely.
The probability measures we derive remain independent of any discretization schemes, and they have infinite degrees of freedom.
Specifically, we will be able to derive a true representation of the posterior over the entire field, rather than a posterior over some approximation of the field.

Suppose we have a field of interest $\phi$ living in a space $\Phi$ that contains functions from a physical domain $\Omega$ to the real numbers.
The appropriate choice of the function space $\Phi$ is problem-specific.
Assume that we have access to a PDE which describes our current state of knowledge about the field
\begin{equation}
    D[\phi] = f,
    \label{eqn:PDE}
\end{equation}
where $D$ is a differential operator and $f$ is the source term.
Particularly, we are interested in PDEs for which the field variables of interest can be obtained through the minimization of a field energy functional $U$ of the form:
\begin{equation}
    U[\phi] = \int_{\Omega}dx\: u\left(x,\phi(x),\nabla_x\phi(x),\dots\right),
    \label{eqn:energy}
\end{equation}
where $u$ is an energy density function which appropriately describes the integrand.
The problem of solving the PDE \qref{PDE} is equivalent to finding the field $\phi$ which minimizes the energy \qref{energy} subject to any boundary conditions.
In many physical systems, a proper energy functional of this form is readily available, e.g., the total energy of the system.
When an energy functional is not available, we can take $U$ to be the integrated squared residual of the PDE:
\begin{equation}
\label{eqn:int_sq_residual}
    U[\phi] = \int_{\Omega}dx\:\left(D[\phi] -f\right)^2. 
\end{equation}

We define the Boltzmann-like probability measure:
\begin{equation}
    p(\phi) \propto \exp\left\{-\beta U[\phi]\right\}.
    \label{eqn:Prior}
\end{equation}
The non-negative scaling parameter $\beta$ plays a role similar to the inverse temperature in statistical field theory.
\qref{Prior} defines a measure which assigns higher probability to functions which are closer to minimizing the energy, \qref{energy} (or square residual \qref{int_sq_residual}).
Additionally, to impose any regularity constraints along with the known physics, the functional prior can be written as 
$$
    p(\phi) \propto \exp\left\{-\beta U[\phi] - \frac{1}{2}\phi^{\dagger}S\phi\right\},
$$
where $S$ is a continuous, bilinear, positive-definite form coming from a standard Gaussian covariance kernel, i.e., the typical inverse covariance operator of classical IFT.

\subsubsection{Modeling the likelihood}
The second ingredient in Bayesian inference is a model of the measurement process, which then enters the likelihood function.
The likelihood connects the underlying physical field to the experimentally measured data.
We introduce the operator $R$ to be a measurement operator, i.e., an $s$-dimensional tuple of functionals.
Physically, the vector $R\phi$ contains the $s$ numbers that we would measure if the underlying physical field was $\phi$ and there was no measurement uncertainty.
In general, we write the likelihood of the data $d$ given the field $\phi$ as
$$
    p(d|\phi) = p(d|R\phi) \propto \exp\left\{-\ell(d,R\phi)\right\},
$$
where $\ell(d,R\phi)$ is a real function equal to the minus log-likelihood up to an additive constant.

If the measurements are independent conditional on the field, then the minus log-likelihood is additive:
\begin{equation}
\label{eqn:like_add}
\ell(d,R\phi) = \sum_{j=1}^s\ell_j(d_j,R_j\phi).
\end{equation}
Usually, the measurement operator samples the field at $s$ specific points $x_j$ in the spatial domain $\Omega$ and the measurement noise can be assumed to be independent at each location with variance $\sigma^2$.
Then, we have:
\begin{equation}
\label{eqn:like_gauss}
\ell(d,R\phi) = \sum_{j=1}^s\frac{(\phi(x_j)-d_j)^2}{2\sigma^2}.
\end{equation}

\subsubsection{The field posterior}
\label{sec:posteriors}
Applying Bayes's rule, the posterior over the field is:
\begin{equation}
    p(\phi |  d) \propto p(d | \phi)p(\phi).
    \label{eqn:posterior}
\end{equation}
This posterior is independent of any parameterization of the field, which is the biggest advantage gained by approaching inference through IFT.
Previous approaches usually are defined in a way which is dependent on the specific spatial/field discretizations, e.g., first using a neural network to learn the functional prior as in \cite{meng2021learning}.
Because of this, the results from these methods depend on the chosen discretization and resolution.
Furthermore, the typical approaches derive posteriors over an \emph{approximation} of the field, rather than the field itself.
With PIFT, we derive a posterior over the field, as seen in \qref{posterior}.
This approach even allows for analytical representations of the posterior under certain circumstances, as seen in \sref{sec:analytic}

The information Hamiltonian corresponding to \qref{posterior} is:
$$
H[\phi|d] = \ell(d,R\phi) + \beta U[\phi].
$$
Observe how the scaling parameter $\beta$ controls our belief in the physics.
When our belief in the physics is strong, we can select a larger value of $\beta$ to increase the effect of the physics on $H[\phi|d]$.
Treating $\beta$ as a hyper-parameter will allow the model to automatically scale this contribution from the physics.
In general, the field seeks a minimum energy state, i.e. the minimum of $H[\phi|d]$ with respect to $\phi$, which corresponds to the maximum probability of $p(\phi|d)$.
This fact further justifies our choice of the physics-informed functional prior.

Once the posterior for the specific problem is derived, the problem then becomes about understanding characteristics of the field from its posterior.
For example, we may be interested in visualizing samples from the posterior field, which will require some numerical scheme.
The PIFT approach is compatible with existing numerical schemes in a way which naturally adds a layer of uncertainty quantification about the field of interest.
We can show that in certain limiting cases, the PIFT approach reduces to previous modeling approaches in a similar manner in which free theory reduces to Gaussian process regression as demonstrated in \sref{sec:ift}.
A numerical scheme for approximating the posterior field based on stochastic gradient Langevin dynamics \cite{welling2011bayesian} is demonstrated in \sref{sec:numericalposts}, but first, we provide an example where an analytic form of the posterior is available.

\subsection{Analytic example: The time-independent Klein-Gordon equation}
\label{sec:analytic}
If the operator which describes the physics is quadratic, we remain in the free theory case.
That means an analytic form of the posterior can be derived in a way similar to the results shown in \sref{free-posterior} for certain problems.
To derive such an example, let us study the time-independent Klein-Gordon equation, otherwise known as the screened Poisson equation:
$$
\left(-\nabla^2 + \alpha^2\right)\phi = q,
$$
in $\R^k$, where $\alpha$ is a positive parameter.
Note that the operator which appears in the Klein-Gordon equation is positive-definite \cite{nakic2020perturbation}, so the various definitions and techniques used in \sref{sec:ift} apply.
The appropriate Hilbert space is the Sobolev space of square integrable functions with square integrable derivatives that vanish sufficiently fast at infinity:
$$
\Phi = \left\{\phi:\R^k\mapsto \R \middle| \int_{\mathbb{R}^k} dx\: \phi^2 + \int_{\mathbb{R}^k} dx\: |\nabla \phi|^2 < \infty, \lim_{R\rightarrow\infty}\int_{B(0,R)}\phi\nabla\phi\cdot n=0\right\},
$$
where $B(0,R)$ represents a ball with radius $R$ in $\mathbb{R}^k$.

It can be shown~\cite{karumuri2019simulator} that solution of the equation makes the following functional stationary:
$$
U[\phi] = \frac{1}{2}\left[(\nabla \phi)^\dagger \nabla \phi + \alpha^2\phi^\dagger\phi \right] - q^\dagger \phi.
$$
Furthermore, the stationary point is a minimum if the boundary conditions are kept fixed.
This suggests the appropriate form for the functional prior is indeed:
\begin{equation}
  p(\phi) \propto \exp\left\{-\beta U[\phi]\right\},  
  \label{eqn:KLprior}
\end{equation}
as the function that solves a given boundary value problem becomes the most probable one.

To make analytical progress, we have to express \qref{KLprior} as a Gaussian prior over fields.
Define the diagonal operator:
$$
S(x,x') = \delta(x-x')\left(-\nabla_{x'}^2 + \alpha^2\right),
$$
where $\nabla_{x'}$ is the gradient with respect to $x'$.
We will show that
$$
U[\phi] = \frac{1}{2}\phi^\dagger S \phi - q^\dagger \phi.
$$
Start by using integration by parts on the first additive term of $U[\phi]$:
$$
\begin{aligned}
(\nabla \phi)^\dagger \nabla \phi &= \int_{\mathbb{R}^k}dx\: \nabla \phi \cdot \nabla \phi \\
&= \lim_{R\rightarrow\infty}\int_{B(0,R)}dx\: \phi \nabla \phi \cdot n - \int_{\mathbb{R}^k}dx\:\phi\nabla^2 \phi.
\end{aligned}
$$
Notice that limit on the second line vanishes by the choice of the function space $\Phi$.
Using the properties of Dirac's delta, the remaining term is the same as $\int_{\R^k\times\R^k}dxdx'\:\phi(x)\delta(x-x')\nabla^2_{x'}\phi(x')$.
This concludes the proof.

We now show that the inverse operator $S^{-1}$ is the Green's function of the Klein-Gordon equation.
The defining equation $SS^{-1}\phi=\phi$ yields:
$$
\int_{\R^k\times \R^{k}}\: dx'dx''\:S(x,x')S^{-1}(x',x'')\phi(x'') = \phi(x).
$$
Using Fubini's theorem, we transform it to:
$$
\int_{\R^k}\:dx''\left[\int_{\R^{k}}\: dx'\:S(x,x')S^{-1}(x',x'')\right]\phi(x'') = \phi(x),
$$
from which we observe that:
$$
\int_{\R^{k}}\: dx'\:S(x,x')S^{-1}(x',x'') = \delta(x-x'').
$$
Plugging in the expression for $S(x,x')$ and integrating over $x'$ gives:
\begin{equation}
\label{eqn:green-eq}
(-\nabla^2+\alpha^2)S^{-1}(x,x'') = \delta(x-x''),
\end{equation}
which proves our claim.

Having established the existence of $S^{-1}$, we can now complete the square:
$$
\begin{aligned}
U[\phi] &= \frac{1}{2}\phi^\dagger S \phi - q^\dagger \phi\\
&= \frac{1}{2}(\phi - S^{-1}q)^\dagger S(\phi - S^{-1}q) + \text{const.}
\end{aligned}
$$
Thus, we have reached our goal of showing that the prior measure is Gaussian.
We write:
$$
p(\phi)\propto \exp\left\{-\frac{\beta}{2}(\phi - S^{-1}q)^\dagger S (\phi - S^{-1}q)\right\},
$$
or
$$
\phi \sim \mathcal{N}(S^{-1}q, \beta^{-1} S^{-1}).
$$
Observe that the mean of the field is:
$$
m = \mathbb{E}[\phi] = S^{-1}q,
$$
which is a solution of the time-independent Klein-Gordon equation.
The covariance of the field is:
$$
\mathbb{C}\left[\phi(x),\phi(x')\right] = \mathbb{E}[\left(\phi(x)-m(x)\right)\left(\phi(x')-m(x')\right)] = \beta^{-1}S^{-1}(x,x').
$$
We see that the covariance is zero when $\beta\rightarrow \infty$, i.e., when we have absolute trust in the physics.
Likewise, when $\beta\rightarrow0$, the covariance approaches infinity, which is true when we do not believe the physics to be true.
This shows that in the limiting behavior the theory is justified.

Let us derive the analytical form of the Green's function $S^{-1}(x,x')$.
By translational symmetry the Green's function is symmetric, $S^{-1}(x,x')=S^{-1}(x-x')$. 
To obtain $S^{-1}$, we apply a Fourier transform on \qref{green-eq}, which results in
$$
S^{-1}(x,x') = \int \frac{d^k\xi}{(2\pi)^k}\frac{\exp\{-i\xi\cdot (x-x')\}}{\xi^2+\alpha^2}.
$$
It can be shown that for a covariance of this form, the corresponding partition function goes to infinity when we consider $\Phi$ to be arbitrarily expressive \cite{tong2017statistical}.
The intuitive reason for this is that the theory considers arbitrarily small scales, and the function space needs to be sufficiently expressive to capture this.
However, this is a common result in statistical field theories, so we do not need to be alarmed.
A common trick to avoid this problem is to restrict $\Phi$ to the smallest possible physical scale of interest, which removes this infinity \cite{lancaster2014quantum}.

Notice that the field prior is not necessarily unimodal.
There may be many fields that maximize the prior probability.
In this particular problem, the prior becomes unimodal only if the candidate fields are constrained to satisfy certain boundary conditions.

Now assume that we observe an $s$-dimensional dataset $d$, through a measurement operator $R$, with additive, zero-mean, Gaussian noise $n$ with covariance matrix $N$.
The posterior over the field becomes:
$$
\begin{aligned}
p(\phi | d) &\propto \mathcal{N}(d|R\phi, N)p(\phi)\\
&= \exp\left\{-\frac{1}{2}(d - R\phi)^\dagger N^{-1}(d - R\phi) -\frac{\beta}{2}(\phi - S^{-1}q)^\dagger S (\phi - S^{-1}q) \right\}.
\end{aligned}
$$
This is a Gaussian random field and we can find its mean and covariance by completing the square.
The posterior inverse covariance operator (scaled by $\beta$) is:
$$
\tilde{S} = S + \beta^{-1}R^\dagger N^{-1}R,
$$
and the posterior mean is:
$$
\tilde{m} = m + \beta^{-1}\tilde{S}^{-1}R^\dagger N^{-1}(d - Rm).
$$
We can find the posterior covariance operator by applying the Sherman-Morrison-Woodbury formula \cite{deng2011generalization}, and it is given by
$$
\tilde{S}^{-1} = S^{-1} - \beta^{-1}S^{-1}R^\dagger K^{-1} RS^{-1},
$$
where 
$$
K= N + \beta^{-1}RS^{-1}R^{\dagger}
$$
is an $s\times s$ positive definite matrix given by 
$$
K_{ij} = N_{ij} + \beta^{-1}r_i^\dagger S^{-1} r_j =  N_{ij} + \beta^{-1}\int dx dx'\: r_i(x) S^{-1}(x,x') r_j(x').
$$
We have constructed an analytic form of the posterior, which is a Gaussian random field given by

\begin{equation}
    p(\phi|d) = \mathcal{N}\left(\phi\middle|\tilde{m}, \beta^{-1}\tilde{S}^{-1}\right).
    \label{eqn:analyticalpost}
\end{equation}

In \qref{analyticalpost}, we have derived a posterior over fields, which is one of the biggest advantages gained through the use of IFT.
Previous approaches which are similar to PIFT such as B-PINNs \cite{yang2021b} or in \cite{meng2021learning} infer a discretized version of the field, and they depend on the specific discretization chosen.
However, the PIFT approach remains independent of any spatial or field discretization, and samples from the PIFT posterior in \qref{analyticalpost} are continuous fields.

\subsection{Sampling scheme for forward problems}
\label{sec:numericalposts}

In the case where an analytical representation of the posterior is not available and we would like to sample directly from the posterior \qref{posterior} we will need to do so numerically.
The standard approach in IFT for numerically constructing the posterior is through metric Gaussian variational inference (MGVI) \cite{knollmuller2019metric}.
There is an open-source python library available called Numerical Information Field Theory (NIFTy) for this approach \cite{selig2013nifty}.
However, the theory and the implementation are not applicable to the case where $S$ is an operator.
Furthermore, MGVI is only equipped to deal with posteriors which can be closely approximated by a Gaussian and for sufficiently linear models.
Because the probability distributions defined here are independent of any parameterization, many different numerical approximation schemes can easily be implemented.
We develop a sampling approach to approximate the posterior based on stochastic gradient Langevin dynamics (SGLD)~\cite{welling2011bayesian}.

We start by parameterizing the field $\phi$ by some class of functions $\hat{\phi}(x;\theta)$ with $c$ trainable parameters $\theta$ in $\mathbb{R}^c$.
Typical choices for $\hat{\phi}$ would be a truncated, complete, orthonormal basis in $\Phi$ or a neural network.
This parameterization will determine the smallest physical scale that can be expressed, so it may be wise to choose a basis with well-known convergence properties.
It is known that feed-forward neural networks are universal approximators in the case of infinite width \cite{scarselli1998universal}.
However, the expressiveness of neural networks is not well understood in general, so it may not always be a wise choice for parameterizing $\phi$.
In our numerical examples, we use a truncated Fourier series expansion, a common choice in SFT.

Without loss of generality, we derive SGLD that sample from the posterior over fields.
This is clearly sufficient since, to sample from the prior, one just needs to remove the data term from the information Hamiltonian.
Recall that the information Hamiltonian has the form
$H[\phi|d] = \ell(d,R\phi) + \beta U[\phi]$, where $\ell(d,R\phi)$ is minus log-likelihood and $U[\phi]$ is the energy functional.
Plugging $\hat{\phi}(\cdot;\theta)$ in place of $\phi$ induces an information Hamiltonian for the field parameters $\theta$:
$$
H(\theta|d) = \ell(d,\theta) + \beta U(\theta),
$$
where we have defined the multivariate functions:
$$
\ell(d,\theta) = \ell(d,R\hat{\phi}(\cdot;\theta)),
$$
and
$$
U(\theta) = U[\hat{\phi}(\cdot;\theta)].
$$
Notice that we have written here $H(\theta | d)$ rather than $H[\theta | d]$. 
This is because $H(\theta | d)$ is a finite-dimensional multivariate function and not a functional over fields as with $H[\phi]$.
To avoid introducing more notation, we adopt this change throughout the paper to distinguish between functions over parameters and functionals over fields, following the convention that the inputs to a function specify what type of function we refer to.
The corresponding posterior over the field parameters is:
$$
p(\theta | d) \propto \exp\left\{-H(\theta|d)\right\}.
$$
We will construct a sampling scheme for $p(\theta|d)$.

SGLD can sample from the field parameter posterior, $p(\theta|d)$, using only noisy, albeit unbiased, estimates of the gradient of the information Hamiltonian, $H(\theta|d)$.
This is very convenient since it allows us to avoid doing the integrals required for the evaluation of the energy of the field, see~\qref{energy}.
Assume for now that we can evaluate the data term $\nabla_{\theta}\ell(d,\theta)$ exactly.
We will construct an unbiased estimator of $U(\theta)$.

Replace the field in \qref{energy} with its parameterization to get:
$$
U(\theta) = \int_{\Omega}dx\:u\left(x,\hat{\phi}(x;\theta),\nabla_x \hat{\phi}(x;\theta),\dots\right).
$$
Introduce an arbitrary random variable $X$ with probability density $q(x)$ with support that covers the physical domain $\Omega$.
In our numerical examples the random variable $X$ is a uniformly distributed in $\Omega$ and $q(x)$ equals the inverse of the volume of $\Omega$.
Dividing and multiplying the integrand by $q(x)$ (importance sampling) we can write $U(\theta)$ as an expectation over $X$:
$$
U(\theta) = \mathbb{E}\left[\frac{u\left(X,\hat{\phi}(X;\theta),\nabla_x \hat{\phi}(X;\theta),\dots\right)}{q(X)}\right].
$$
To construct the desired unbiased estimator, take $X_1,\dots,X_n$ to be independent identically distributed (iid) random variables following $q$ and observe that:
$$
\nabla_\theta U(\theta) = \frac{1}{n}\sum_{i=1}^n \mathbb{E}\left[\frac{\nabla_\theta u\left(X_i,\hat{\phi}(X_i;\theta),\nabla_x \hat{\phi}(X_i;\theta),\dots\right)}{q(X_i)}\right].
$$

We are now ready to state the SGLD for $\theta$.
We start with an arbitrary $\theta_1$ and iterate as:

\begin{equation}
    \label{eqn:SGLDforward}
    \theta_{t+1} = \theta_t - \epsilon_t \left[\nabla_\theta \ell(d,\theta) + \frac{1}{n}\sum_{i=1}^n \frac{\nabla_\theta u\left(x_{ti},\hat{\phi}(x_{ti};\theta_t),\nabla_x \hat{\phi}(x_{ti};\theta_t),\dots\right)}{q(x_{ti})}\right] + \eta_t,
\end{equation}
where the spatial points $x_{ti}$ are all independent samples from the importance sampling distribution $q$, $\epsilon_t$ is the learning rate series, and $\eta_t$ is a random disturbance.
As shown in \cite{welling2011bayesian} the algorithm produces a Markov chain that samples from the desired distribution if the learning rate $\epsilon_t$ satisfies the Robbins-Monro conditions~\cite{robbins1951stochastic}:
$$
\sum_{t=1}^\infty \epsilon_t = +\infty,\;\sum_{t=1}^\infty\epsilon_t^2 < +\infty,
$$
and the random disturbances are iid with:
$$
\eta_t \sim \mathcal{N}(0, \sqrt{\epsilon_t}).
$$
The learning rate series we use throughout the numerical examples is:
$$
\epsilon_t = \frac{\alpha_0}{t^{\alpha_1}},
$$
with $\alpha_0$ positive and $\alpha_1$ in $(0.5,1]$.

An algorithm for sampling the PIFT prior using the SGLD scheme is derived in \qref{SGLDforward} can be seen in \aref{priorsamps}.
As we will see later, samples from the prior are necessary for the scheme implemented to solve inverse problems.
An algorithm for sampling the posterior is reserved for the next section, which subsamples the data.

\begin{algorithm}
\caption{SGLD for PIFT priors}\label{alg:priorsamps}
\begin{algorithmic}[1]
\State{\textbf{Input:} $\theta_1$, $\alpha_0$, $\alpha_1$, $n$, $k$}
\For{$t=1,\dots,k$}
    \State Set $\epsilon_t \leftarrow \frac{\alpha_0}{t^{\alpha_1}}$.
    \State Sample $\eta_t$ from $\mathcal{N}(0,\sqrt{\epsilon_t})$.
    \State Sample $n$ $x_{ti}$ from $q(x_{ti})$.
    \State Set $\theta_{t+1} \leftarrow \theta_t - \frac{\epsilon_t}{n}\sum_{i=1}^n\frac{\nabla_{\theta}u\left(x_{ti},\hat{\phi}(x_{ti};\theta_t),\nabla_x\hat{\phi}(x_{ti};\theta_t),\dots\right)}{q(x_{ti})} + \eta_t$.
\EndFor
\State \textbf{Output:} $\:\theta_1,\dots,\theta_k$
\end{algorithmic}
\end{algorithm}

\subsubsection{Scaling to a large number of observations}
Let us show how one can construct an algorithm that is scalable to large numbers of observations if the observations are independent.
In this case, the data term $\ell(d,\theta)$ is additive, see~\qref{like_add}:
$$
\ell(d,\theta) = \sum_{j=1}^s\ell_j(d,\theta),
$$
with $\ell_j(d,\theta) = \ell_j(d,R\phi(\cdot;\theta))$.
We can construct an unbiased estimator of the data term by introducing a categorical random variable $J$ taking values from $1$ to $s$ with probability $1/s$ and noticing that:
$$
\ell(d,\theta) = s\mathbb{E}\left[\ell_{J}(d,\theta)\right].
$$
Notice the multiplication by the number of samples $s$.
The modified SGLD for theta are:
\begin{equation}
\label{eqn:sgld}
\theta_{t+1} = \theta_t - \epsilon_t \left[\frac{s}{b}\sum_{i=1}^b\nabla_\theta \ell_{j_{ti}}(d,\theta) + \frac{1}{n}\sum_{i=1}^n \frac{\nabla_\theta u\left(x_{ti},\hat{\phi}(x_{ti};\theta_t),\nabla_x \hat{\phi}(x_{ti};\theta_t),\dots\right)}{q(x_{ti})}\right] + \eta_t,
\end{equation}
where $b$ is a positive integer, and $j_{ti}$ are iid samples from $J$, and the rest of the variables are as in the previous paragraph.

An algorithm for sampling the PIFT posterior using the SGLD scheme with data subsampling derived in \qref{sgld} can be seen in \aref{postsamps}. 

\begin{algorithm}
\caption{SGLD for PIFT posteriors}\label{alg:postsamps}
\begin{algorithmic}[1]
\State{\textbf{Input:} $\theta_1$, $\alpha_0$, $\alpha_1$, $n$, $b$, $k$}
\For{$t=1,\dots,k$}
    \State Set $\epsilon_t \leftarrow \frac{\alpha_0}{t^{\alpha_1}}$.
    \State Sample $\eta_t$ from $\mathcal{N}(0,\sqrt{\epsilon_t})$.
    \State Sample $n$ $x_{ti}$ from $q(x_{ti})$.
    \State Collect $b$ $j_{ti}$ samples from $J$.
    \State Set $\theta_{t+1} \leftarrow \theta_t - \epsilon_t \left[\frac{s}{b}\sum_{i=1}^b\nabla_\theta \ell_{j_{ti}}(d,\theta) + \frac{1}{n}\sum_{i=1}^n \frac{\nabla_\theta u\left(x_{ti},\hat{\phi}(x_{ti};\theta_t),\nabla_x \hat{\phi}(x_{ti};\theta_t),\dots\right)}{q(x_{ti})}\right] + \eta_t$.
\EndFor
\State \textbf{Output:} $\:\theta_1,\dots,\theta_k$
\end{algorithmic}
\end{algorithm}

\subsubsection{On the choice of hyper-parameters}
\label{sec:sgld_params_1}
In all of our numerical examples, except the last one, we use SGLD with the following hyper-parameters.
We choose the number of spatial points to be 1 point and the data batch size we choose $n=1$ and $b=1$, respectively.
For the learning rate decay, we choose $\alpha_1 = 0.51$.
The choice of the learning rate $\alpha_0$ is more critical.
If the learning rate is too small, then the algorithm requires an exceedingly large number of iterations.
If it is too big, then there are numerical instabilities.
In our numerical examples, the following choices work well.
For prior sampling, we always use $\alpha_0 = \frac{0.1}{\beta}$.
For posterior sampling, we use $\alpha_0 = \frac{\hat{\alpha}_0}{\max\left\{\beta,\sigma^{-2}\right\}}$, where $\sigma^2$ is the measurement noise variance, see~\qref{like_gauss}.
The value of $\hat{\alpha}_0$ is reported in each numerical experiment.

\subsubsection{Sampling multi-modal field posteriors}
SGLD is not suitable for multi-modal distributions.
In our last numerical example, in which the field posterior is bimodal, we use a more advanced method based on
a variant of the Hamiltonian Monte Carlo (HMC) algorithm \cite{betancourt2017conceptual}.
HMC is a Markov chain Monte Carlo \cite{neal2011mcmc} algorithm that is designed for sampling from distributions with high dimensionality.
Specifically, we implement the HMC with energy conserving subsampling (HMCECS) \cite{dang2019hamiltonian}, which is an extension of the stochastic gradient Hamiltonian Monte Carlo (SGHMC) \cite{chen2014stochastic} that remains invariant to the desired target distribution when some type of subsampling scheme is used.
We take advantage of this property of HMCECS to evaluate the spatial integral coming from the energy functional with a subsampling scheme to improve the efficiency of the algorithm.

For users of the algorithms in this paper, a method which can capture multi-modal posteriors such as HMC should be preferred unless the posterior is strongly suspected to be unimodal.
For example, if the energy functional is strictly convex, then SGLD is appropriate and simpler to implement.
More research is needed to provide further insight into the nature of these algorithms.

\subsection{Inverse problems}
\label{inverse}
Consider the case of inverse problems where there are parameters of the likelihood, the energy functional, or the boundary conditions that need to be inferred alongside the field.
For any given parameter, if this parameter does not impact the partition function, e.g., any parameters of a Gaussian likelihood, then this parameter can be sampled with the field using SGLD as in \sref{sec:numericalposts}.
However, if the partition function depends on a parameter, e.g., the scale parameter $\beta$ or any parameter of the energy functional, the SGLD scheme we developed is not applicable.
The purpose of this section is to develop an approach that deals with the latter case.

Let $\lambda$ denote the unknown parameters of the energy functional to be inferred.
Note that $\lambda$ may include fields which appear in the energy functional as PIFT is equipped to handle inference over fields in general. In our numerical examples we infer a source term of a non-linear PDE.
Let $H[\lambda, \beta]$ be the prior information Hamiltonian for $\lambda$ and $\beta$, i.e.,
$$
H[\lambda, \beta] = -\log p(\lambda, \beta).
$$
The exact choice of $p(\lambda, \beta)$ is problem-specific and should be selected to be consistent to the current knowledge of the physics, e.g., selecting probability densities with positive support over parameters which cannot be negative.
For notational convenience, introduce the prior information Hamiltonian for the physical field $\phi$ conditional on the parameters $\lambda$ and $\beta$:
$$
H[\phi|\lambda, \beta] = \beta U[\phi|\lambda].
$$
The field prior conditional on the parameters is:
$$
p(\phi|\lambda, \beta) = \frac{\exp\left\{-H[\phi|\lambda, \beta]\right\}}{Z[\lambda, \beta]},
$$
with a partition function:
$$
Z[\lambda,\beta] = \int_{\Phi}\mathcal{D}\phi\:\exp\left\{-H[\phi|\lambda,\beta]\right\}.
$$

We show that it is incorrect to ignore the partition function when inferring $\lambda$ or $\beta$.
Assume that we have measurements $d$ and a data likelihood $p(d|\phi)$.
Applying Bayes's rule, the joint posterior over the parameters $\lambda$ and $\beta$ and the field $\phi$ is:
\begin{equation}
\label{eqn:joint}
p(\phi,\lambda, \beta|d) = \frac{p(d|\phi)p(\phi|\lambda, \beta)p(\lambda, \beta)}{p(d)} = \frac{1}{p(d)}p(d|\phi)p(\lambda,\beta)\frac{\exp\left\{-H[\phi|\lambda, \beta]\right\}}{Z[\lambda,\beta].
}
\end{equation}
Our state of knowledge about the parameters $\lambda$ and $\beta$ after seeing the data, is neatly captured by the marginal:
\begin{equation}
\label{eqn:lambda_marginal}
p(\lambda, \beta|d) = \int_{\Phi}\mathcal{D}\phi\: p(\phi,\lambda, \beta|d) = \frac{1}{p(d)}\int_{\Phi}\mathcal{D}\phi\:p(d|\phi)p(\lambda,\beta)\frac{\exp\left\{-H[\phi|\lambda, \beta])\right\}}{Z[\lambda, \beta]}.
\end{equation}
This marginal depends on $Z[\lambda, \beta]$, so the latter cannot be ignored.
This implies that algorithms which directly sample the posterior are not appropriate unless the partition function can be approximated.
Rather than developing an approximation for the partition function compatible with HMCECS, we instead derive a nested SGLD approach, as described in the next section.
This nested approach allows the partition function to be dropped altogether.

\subsubsection{Sampling scheme for inverse problems}
\label{sec:inversesampling}
We derive a nested SGLD scheme that samples from the marginal posterior of the parameters $\lambda$ and $\beta$.
Note that from this point forward we absorb $\beta$ into $\lambda$ to simplify the notation, as the mathematics remain the same.
The nested SGLD approach contains an inner loop of SGLD where samples are generated from the field posterior and prior for a fixed set of parameters $\lambda_t$.
This requires solving a forward problem, and the partition function may be dropped as described in \sref{sec:numericalposts}.
Then, the parameters $\lambda_t$ are updated in an outer loop of SGLD using the samples of the field obtained from the forward problem.

The key ingredient for the nested SGLD approach is the derivative of the information Hamiltonian with respect to the parameters $\lambda$.
We will derive a formula for this derivative using properties of expectations.
To proceed, we first need to parameterize any fields in $\lambda$ using a finite set of numbers.
To avoid adding more notation, please think of $\lambda$ in this section as a finite dimensional vector.
That is, if $\lambda$ contained a field, simply replace that field with a finite dimensional parameterization of said field.
To reinforce this aspect, we will be writing $H(\lambda)$ instead of $H[\lambda]$
and $H(\lambda|d)$ instead of $H[\lambda|d]$.
That is, if $\lambda$ is finite-dimensional these information Hamiltonians are multivariate functions instead of functionals.

Let $H(\lambda|d)$ be the information Hamiltonian corresponding to the marginal posterior of \qref{lambda_marginal}, i.e.,
$$
H(\lambda|d) = -\log p(\lambda|d).
$$
SGLD requires an unbiased estimator of gradient of the information Hamiltonian with respect to $\lambda$.
Fortunately, this is possible without having to estimate the partition function.
Specifically, we will derive an unbiased estimator that uses only samples from the prior and the posterior of the physical field $\phi$.
So, we seek an unbiased estimator of the gradient  $\nabla_{\lambda}H(\lambda | d)$, which can be found using properties of expectations.

Let $\mathbb{E}[\cdot|\lambda]$ and $\mathbb{E}[\cdot|d,\lambda]$ denote the conditional expectation operators with respect to the physical field $\phi$ over the prior and posterior probability measures, respectively.
For example, if $F$ is a functional,
$$
\mathbb{E}\left[F[\phi]\middle|\lambda\right] = \int_{\Phi}\mathcal{D}\phi\:p(\phi|\lambda)F[\phi]. 
$$
We assert that the gradient of the information Hamiltonian $H(\lambda|d)$ with respect to $\lambda$ is given by the difference of the posterior expectation minus the prior expectation of the gradient of the information Hamiltonian of the field, i.e.:
\begin{equation}
\label{eqn:unbiased_grad_lambda}
\nabla_{\lambda}H(\lambda|d) = \mathbb{E}\left[\nabla_{\lambda}H[\phi|\lambda]\middle| d,\lambda\right] - \mathbb{E}\left[\nabla_{\lambda}H[\phi|\lambda]\middle| \lambda \right]  + \nabla_{\lambda} H(\lambda) .
\end{equation}
The derivation is long, but uses only standard differentiation rules, see Appendix \ref{appendix:unbiased_grad_lambda_derivation}.

Progressing towards the desired unbiased estimator, we express $\nabla_{\lambda}H(\lambda|d)$ as a single expectation.
Recall that the information Hamiltonian is $H[\phi|\lambda]=\beta U[\phi|\lambda]$ and that the energy $U[\phi|\lambda]$ is the integral over the domain $\Omega$ of
some energy density, $u\left(x,\phi(x),\nabla_x\phi(x),\dots;\lambda\right)$, which may depend on $\lambda$.
So, we can write:
$$
H[\phi|\lambda] = \int_\Omega dx\:h(x,\phi(x),\nabla_x\phi(x),\dots;\lambda),
$$
where, to keep the notation unified, we have defined the Hamiltonian density:
$$
h(x,\phi(x),\nabla_x\phi(x),\dots;\lambda) = \beta u(x,\phi(x),\nabla_x\phi(x),\dots;\lambda).
$$
As in \sref{sec:numericalposts}, we introduce an importance sampling distribution $q$ and a random variable $X$ following $q$.
We can now write:
$$
H[\phi|\lambda] = \mathbb{E}\left[\frac{h(X,\phi(X),\nabla_x\phi(X),\dots;\lambda)}{q(X)}\middle| \phi, \lambda \right].
$$
Putting everything together, we have shown that:
\begin{align*}
\nabla_{\lambda}H(\lambda|d) =& \mathbb{E}\left[\frac{\nabla_{\lambda}h(X,\phi(X),\nabla_x\phi(X),\dots;\lambda)}{q(X)}\middle| d,\lambda\right]\\
&- \mathbb{E}\left[\frac{\nabla_{\lambda}h(X,\phi(X),\nabla_x\phi(X),\dots;\lambda)}{q(X)}\middle| \lambda \right]\\
&+ \nabla_{\lambda} H(\lambda),
\end{align*}
where the expectations are over all the random objects on which we do not condition.

We are now ready to state the nested SGLD scheme that samples from the marginal posterior of $\lambda$.
Let $\phi_i$ and $\tilde{\phi}_i$ be $k$ and $\tilde{k}$ samples of the field $\phi$ from the prior and the posterior 
conditional on $\lambda$, respectively.
Of course, we are actually sampling finite-dimensional parameterization of the fields.
We produce these samples by doing $T$ and $\tilde{T}$ steps of the SGLD scheme presented in~\sref{sec:numericalposts}.
Let $x_{tj}$ and $\tilde{x}_{tj}$ be $n$ and $\tilde{n}$ independent samples from the importance sampling distribution $q$.
Starting from an initial $\lambda_1$, we iterate as:
\begin{equation}
    \label{eqn:SGLDinverse}
    \begin{aligned}
    \lambda_{t+1} =& \lambda_t\\
    &-\epsilon_t \Bigg[\frac{1}{\tilde{k}\tilde{n}}\sum_{i=1}^{\tilde{k}}\sum_{j=1}^{\tilde{n}}\frac{\nabla_{\lambda}h(\tilde{x}_{tj},\tilde{\phi}_{ti}(\tilde{x}_{tj}),\nabla_x\tilde{\phi}_{ti}(\tilde{x}_{tj}),\dots;\lambda_t)}{q(\tilde{x}_{tj})}\\
    &-
    \frac{1}{kn}\sum_{i=1}^{k}\sum_{j=1}^{n}\frac{\nabla_{\lambda}h(x_{tj},\phi_{ti}(x_{tj}),\nabla_x\phi_{ti}(x_{tj}),\dots;\lambda_t)}{q(x_{tj})}\\
    &+\nabla_{\lambda}H(\lambda_t)
    \Bigg]\\
    &+\eta_t,
    \end{aligned}
\end{equation}
where the learning rate $\epsilon_t$ and the random disturbance $\eta_t$ satisfy the same conditions as in~\sref{sec:numericalposts}.

An algorithm which neatly descibes the steps needed to evaluate \qref{SGLDinverse} can be seen in \aref{inverse}.
As we can see, this algorithm can make use of the previous algorithms already defined for forward problems and clearly shows how we have derived a nested SLGD scheme for moving the parameters.

\begin{algorithm}
\caption{SGLD for PIFT inverse problems}\label{alg:inverse}
\begin{algorithmic}[1]
\State{\textbf{Input:} $\lambda_0$ (initial parameters), $\alpha_0$, $\alpha_1$ (learning rate), $b$, $n$, $\tilde{n}$, $T$, $\tilde{T}$, $k$, $\tilde{k}$, $\text{maxiter}$}
\For{$t=1,\dots,\text{maxiter}$}
    \State $\epsilon_t \leftarrow \frac{\alpha_0}{t^{\alpha_1}}$.
    \State Sample $\eta_t$ from $\mathcal{N}(0,\sqrt{\epsilon_t})$.
    \State Select $k$ $\phi_i$ from $T$ prior samples from \aref{priorsamps} with $\lambda_t$ fixed, $n$ $x_{tj}$ sampled from $q(x_{tj})$.
    \State Select $\tilde{k}$ $\tilde{\phi}_i$ from $\tilde{T}$ posterior samples from \aref{postsamps} with $\lambda_t$ fixed, $\tilde{n}$ $\tilde{x}_{tj}$ sampled from $q(\tilde{x}_{tj})$.
    \State Update $\lambda_t$ according to \qref{SGLDinverse}.
\EndFor
\State \textbf{Output:} $\:\lambda_1,\dots,\lambda_{\text{maxiter}}$ (samples from posterior of parameters)
\end{algorithmic}
\end{algorithm}

\subsubsection{On the choice of hyper-parameters}
\label{sec:sgld_params_2}
In all our numerical examples, we do one million steps of prior and posterior field sampling at $\lambda=\lambda_1$ before we start updating $\lambda$.
Then, for each iteration $t$, we perform $T=10$ prior SGLD iterations and $\tilde{T}=1$ posterior SGLD iterations.
The choice $\tilde{T}=1$ is justified empirically by the fact that, in our numerical examples, the posterior of $\phi$ is not sensitive on  parameters being calibrated.
Note that this would not have been a good choice if the parameters affected the likelihood.
We choose $n=\tilde{n}=1$ and $q$ to be the uniform on $\Omega$.
The learning rate decay is set to $\alpha_1 = 0.51$.
The learning rate $a_0$ depends on the scaling of the numerical example and we report it later.

\subsection{Relation to physics-informed neural networks}
\label{sec:relation_to_pinns}

\subsubsection{Relation to classical physics-informed neural networks}
\paragraph{Classical physics-informed neural networks are equivalent to maximum a posteriori field estimates.}
Suppose we have a differential equation $D[\phi]=f$ in a domain $\Omega$.
In addition, we have $s$ noisy measurements of the field at spatial points $x_j$ in $\Omega$.
As before, we use $d$ to collectively denote the measurements.
In the case of classical PINNs \cite{raissi2017physics}, we parameterize the field using a neural network $\hat{\phi}(\cdot;\theta)$, where $\theta$ are the network parameters.
A physics-informed loss function is formed by combining residuals of the PDE and measurements,
\begin{equation}
    L(\theta) = \frac{r_d}{s}\sum_{i=1}^s\left[d_i-\hat{\phi}(x_i;\theta)\right]^2 + r_p\int_{\Omega}dx\left[D[\hat{\phi}(x;\theta)](x)-f(x)\right]^2,
    \label{eqn:pinns-loss}
\end{equation}
where $r_d$ and $r_p$ are additional weighting parameters for the data and physics, respectively.
Typically, one approximates the integral in the equation above using a sampling average.
Then, the problem of solving the PDE becomes an optimization problem, where the solution is found by minimizing $L(\theta)$ with respect to $\theta$.
Typically, this problem is cast as a $\emph{stochastic optimization problem}$, where the integral in \qref{pinns-loss} is rewritten as an expectation, and a stochastic gradient descent algorithm such as ADAM is used \cite{kingma2014adam}.
For the specific details, see \cite{karumuri2019simulator}.

We show that, when the field is parameterized by a neural network, and for suitable choices of the weighting parameters and the energy functional, classical PINNs yield a MAP estimate of the PIFT posterior.
In other words, the loss function $L(\theta)$ is minus the logarithm of the PIFT posterior, i.e., it is equal to the information Hamiltonian of the neural network parameters $\theta$.

To prove our claim, start by picking the field energy to be the integrated squared residual of~\qref{int_sq_residual}.
Then, assume that the measurements are independent and that the measurement noise is zero-mean Gaussian with variance $\sigma^2$.
The minus log-likelihood is given by~\qref{like_gauss}.
Finally, parameterize $\phi$ with the neural network $\hat{\phi}(\cdot;\theta)$.
The information Hamiltonian for the parameters, $\hat{H}(d,\theta)$, is obtained from the field information Hamiltonian $H[d,\phi]$ after replacing the field with its parameterization.
It is:
$$
\hat{H}(d,\theta) = \frac{1}{2\sigma^2}\sum_{j=1}^s\left[\hat{\phi}(x_j;\theta) - d_j\right]^2 + \beta\int_{\Omega}dx\left[D[\hat{\phi}(x;\theta)](x)-f(x)\right]^2.
$$
Direct comparison with \qref{pinns-loss} reveals that the PINNS loss is the same as the parameter information Hamiltonian if $r_d = \frac{s}{2\sigma^2}$ and $r_p = \beta$.

\paragraph{Parameter estimation in classical physics-informed neural networks is equivalent to a maximum a posteriori PIFT estimate if the parameters' prior is flat and the partition function is a constant.}
For concreteness, consider the case in which the differential operator depends on some parameters $\lambda$, i.e., the equation is $D[\phi;\lambda]=f$.
The loss function is now a function of both $\theta$ and $\lambda$,
\begin{equation}
\label{eqn:pinns_inverse}
L(\theta,\lambda) = \frac{r_d}{s}\sum_{i=1}^s\left[d_i-\hat{\phi}(x_i;\theta)\right]^2 + r_p\int_{\Omega}dx\left[D[\hat{\phi}(x;\theta);\lambda](x)-f(x)\right]^2.
\end{equation}
One then proceeds to minimize the loss function over both $\theta$ and $\lambda$ simultaneously.

We show that classical PINNs parameter estimation is equivalent to a simultaneous field and parameter MAP estimate in PIFT, if one uses a flat parameter prior and the partition function does not depend on the parameters.
The negative logarithm of the joint posterior of the field $\theta$ and $\lambda$ can be obtained by replacing the field in~\qref{joint} by its parameterization.
This yields:
\begin{equation}
\label{eqn:pift_joint_post}
-\log p(\theta,\lambda|d) = \frac{1}{2\sigma^2}\sum_{j=1}^s\left[\hat{\phi}(x_j;\theta) - d_j\right]^2 + \beta\int_{\Omega}dx\left[D[\hat{\phi}(x;\theta);\lambda](x)-f(x)\right]^2 + \log Z(\lambda) + H(\lambda).
\end{equation}
Comparing with \qref{pinns_inverse} we observe that the two are the same if $H(\lambda)$ and $\log Z(\lambda)$ are constants.

\subsubsection{Relation to Bayesian physics-informed neural networks}
In B-PINNs~\cite{yang2021b}, one starts by parameterizing the field using a neural network, see previous subsection, albeit they also introduce a prior $p(\theta)$ over the neural network parameters, typically a zero-mean Gaussian.
$$
p(\theta) = N(\theta|0,\sigma_\theta^2I) = (2\pi)^{-\frac{1}{2}}\sigma_\theta^{-1}\exp\left\{-\frac{\theta^2}{2\sigma_\theta^2}\right\}.
$$
Furthermore, one introduces a prior over any unknown parameters $\lambda$, say $p(\lambda) = e^{-H(\lambda)}$.

The physics come into the picture through a fictitious set of observations of the residual of the differential equation.
In particular, one introduces $n$ collocation points $x_i^c$ in the domain $\Omega$ and defines the residuals:
$$
\hat{r}_i(x_i;\theta,\lambda) = D[\hat{\phi}(x_i^c;\theta);\lambda] - f(x_i).
$$
The assumption is that the residuals are random variables, say $R_i$, which normally distributed about $\hat{r}_i(x_i;\theta,\lambda)$ with some noise variance $\sigma_r^2$.
That is:
$$
R_i | \theta, \lambda \sim \mathcal{N}(\hat{r}_i(x_i^c;\theta,\lambda), \sigma^2_r).
$$
One then creates the fictitious dataset by assuming that the ``observed'' residuals, say $r_i$, are identically zero.
Denote by $r=(r_1,\dots,r_n)$ the vector of ``observed'' residuals.
Assuming that the residuals are independent, we get a likelihood term of the form:
$$
p(r=0|\theta,\lambda) = \prod_{i=1}^np(r_i|\theta,\lambda) = \left(2\pi\right)^{-\frac{n}{2}}\sigma_r^{-n}\exp\left\{-\frac{\sum_{i=1}^n\left[D[\hat{\phi}(x_i^c;\theta);\lambda] - f(x_i^c)\right]^2}{2\sigma_r^2}\right\}.
$$

Regular observations $d$ appear in the same way as in PIFT.
In case of point observations of the field with independent Gaussian measurement noise with variance $\sigma^2$ at $s$ locations $x_j$, we have:
$$
p(d|\theta) = (2\pi)^{-\frac{s}{2}}\sigma^2\exp\left\{-\frac{\sum_{j=1}^s\left[\hat{\phi}(x_j;\theta) - d_j\right]^2}{2\sigma^2}\right\}.
$$

Putting everything together, the posterior over the field parameters $\theta$ and the physics parameters $\lambda$ is:
$$
p(\theta,\lambda|d,r=0) \propto p(d|\theta)p(r=0|\theta,\lambda)p(\theta)p(\lambda).
$$
The negative logarithm of this quantity is:
\begin{equation}
\label{eqn:bpinns}
\begin{array}{ccc}
-\log p(\theta,\lambda|d,r=0) &=& \frac{1}{2\sigma^2}\sum_{j=1}^s\left[\hat{\phi}(x_j;\theta) - d_j\right]^2\\
&&+
\frac{1}{2\sigma_r^2}\sum_{i=1}^n\left[D[\hat{\phi}(x_i^c;\theta);\lambda] - f(x_i^c)\right]^2\\
&&+ \frac{1}{2\sigma_\theta^2}\lVert\theta\rVert^2 + H(\lambda) + \text{const.}
\end{array}
\end{equation}

For the same problem, using the integrated squared residual for the energy functional, we have already derived the same quantity in PIFT in~\qref{pift_joint_post}.
If we approximate the spatial integral using a sampling average that utilizes the same collocation points as B-PINNs, then we have the approximation:
$$
\begin{array}{ccc}
-\log p(\theta,\lambda|d) &\approx& \frac{1}{2\sigma^2}\sum_{j=1}^s\left[\hat{\phi}(x_j;\theta) - d_j\right]^2\\
&&+ \beta\frac{1}{n}\sum_{i=1}^{n}\left[D[\hat{\phi}(x_i^c;\theta);\lambda] - f(x_i^c)\right]^2\\
&&+\log Z(\lambda) + H(\lambda).
\end{array}
$$

Comparing the equation above with \qref{bpinns}, we see that, in general, PIFT and B-PINNs do not yield the same joint posterior.
This is obvious for the energy parameters $\lambda$ since B-PINNs does not include the partition function term.
However, the two approaches agree on the field parameter posterior $p(\theta|d,\lambda)$ only if:
(i) we approximate the PIFT spatial integral with the same collocation points as B-PINNs;
(ii) we choose a flat prior over the neural network parameters, i.e., $\sigma_\theta=\infty$; and
(iii) the fictitious residual variance is set to:
\begin{equation}
\label{eqn:sigma_r}
\sigma_r^2 = \frac{n}{2\beta}.
\end{equation}
In PIFT, $\beta$ is a parameter that controls our beliefs about the physics.
In B-PINNS, the residual variance $\sigma_r$ cannot be interpreted in the same way.
To further understand the nuances in assigning such an interpretation to $\sigma_r^2$, recall a result of Bayesian asymptotic analysis~\cite[Appendix B]{gelman2013bayesian}.
Under fairly general assumptions (iid measurements, continuous likelihood as a function of the parameters), if $\sigma_r$ is kept fixed while the number of collocation points $n$ goes to infinity, then the ``posterior distribution of [the field parameters] $\theta$ [conditional on the energy parameters $\lambda$] approaches normality with mean $\theta_0$ and variance $(nJ(\theta_0))^{-1}$,'' where the value $\theta_0$ minimizes the Kullback-Leibler divergence between the model residual distribution and the actual residual distribution (which is a delta function centered at zero), while $J(\theta)$ is the Fisher information.
In particular, notice that as the number of collocation points goes to infinity the posterior uncertainty over $\theta$ goes to zero.
In other words, the field posterior in B-PINNs collapses to a single function irrespective of the value of $\sigma_r$ as $n\rightarrow\infty$.
This asymptotic behavior is not necessarily a problem if one has absolute trust in the physics and the boundary conditions are sufficient to yield a unique solution.
However, a collapsing field posterior is undesirable in cases of model-form uncertainty or in ill-posed problems.
PIFT does not suffer from the same problem.

\section{Numerical Examples}
\label{sec:examples}

We demonstrate how the method performs on linear and nonlinear differential equations. 
The examples are chosen in a way which showcase different characteristics of the method.
Examples 1 to 3 are implemented in C++ for computational efficiency required for the nested SGLD schemes.
Example 4, which uses advanced MCMC sampling albeit in only a forward setting, is implemented in NumPyro \cite{phan2019composable}.
We have published the code that reproduces all the results of the paper in the GitHub repository:
\url{https://github.com/PredictiveScienceLab/pift-paper-2023}.

\subsection{Example 1 -- Empirical evidence that the posterior collapses as inverse temperature increases to infinity}
\label{sec:linearproblem}
The first example we cover is the general form of the 1D steady-state heat equation
\begin{equation}
    D\frac{d^2\phi}{dx^2} + q = 0, \label{eqn:heat-eq}
\end{equation}
for $x$ in $[0,1]$ with boundary conditions
$$
\phi(0) = 1,\;\text{and}\;\phi(1) = 0.1.
$$
The field $\phi$ represents the temperature profile along a thin rod, $D$ is the thermal conductivity, and $q$ is the source term.
We can assume that candidate solutions to \qref{heat-eq} are functions $\phi$ in the Hilbert space $\Phi = L^2([0,1],\mathbb{R})$.
In this example, we empirically demonstrate how the inverse temperature $\beta$ controls the variance of the posterior.
Specifically, we see that as $\beta\rightarrow\infty$, the posterior collapses to a single field.

To begin, we must pose the problem defined in \qref{heat-eq} as an equivalent minimization of some functional.
For this, we can turn to the classic statement of Dirichlet's principle, which transforms the problem of solving a Poisson equation into an appropriate minimization of the Dirichlet energy.
The appropriate energy functional is
\begin{equation}
    U[\phi] = \int_0^1dx\left(\frac{1}{2}D\left|\frac{d\phi}{dx}\right|^2-\phi q \right), \label{eqn:heat-energy}
\end{equation}
where the minimization of \qref{heat-energy} provides the solution to \qref{heat-eq} subject to any boundary conditions \cite{courant2005dirichlet}.

For this example, we select $D = 1/4$.
Additionally, we select the source term to be given by
$$
q(x) = \exp{(-x)}.
$$
We parameterize $\phi$ so that it automatically satisfies the boundary conditions.
Specifically, we write:
\begin{equation}
\label{eqn:boundary_satisfied}
\hat{\phi}(x;\theta) = (1-x)\phi(0) + x\phi(1) + (1-x)x\hat{\psi}(x;\theta),
\end{equation}
and we choose $\hat{\psi}(x;\theta)$ to be a Truncated Fourier expansion on $[0,1]$:
\begin{equation}
\psi(x;\theta) = \theta_1 + \sum_{j=1}^K\left[\theta_{1+j}\cos\left(2\pi j x\right)+\theta_{1+K+j}\sin\left(2\pi j x\right)\right].
\end{equation}
We set $K=20$ which gives at total of $1+2(K-1)=39$ field parameters.

We sample the field parameters using SGLD with the hyper-parameter values specified in~\ref{sec:sgld_params_1}.
The results are shown in \fref{ex1compiled}.
We see that under the same conditions, with larger values of $\beta$ the posterior begins to collapse to a single function.

\begin{figure}[hbt!]
    \centering
    \begin{subfigure}[b]{0.475\textwidth}
        \centering
        \includegraphics[width=\textwidth]{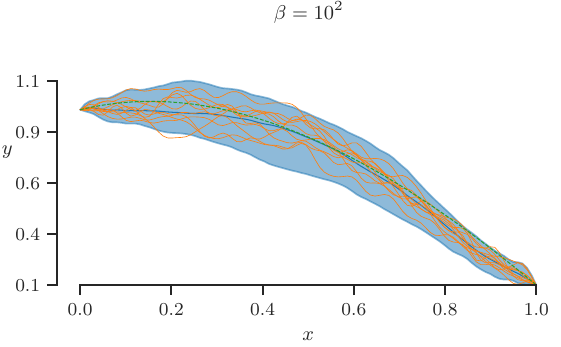}
        \label{fig:ex1a}
    \end{subfigure}
    \hfill
    \begin{subfigure}[b]{0.475\textwidth}
        \centering
        \includegraphics[width=\textwidth]{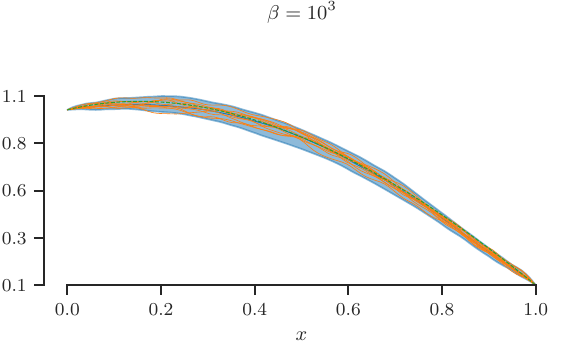}
        \label{fig:ex1b}
    \end{subfigure}
    \vskip\baselineskip
    \begin{subfigure}[b]{0.475\textwidth}
        \centering
        \includegraphics[width=\textwidth]{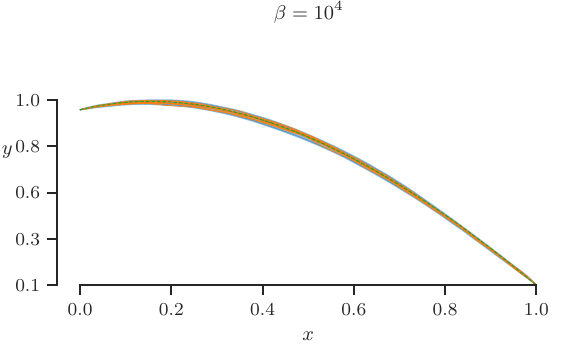}
        \label{fig:ex1c}
    \end{subfigure}
   \hfill
    \begin{subfigure}[b]{0.475\textwidth}
        \centering
        \includegraphics[width=\textwidth]{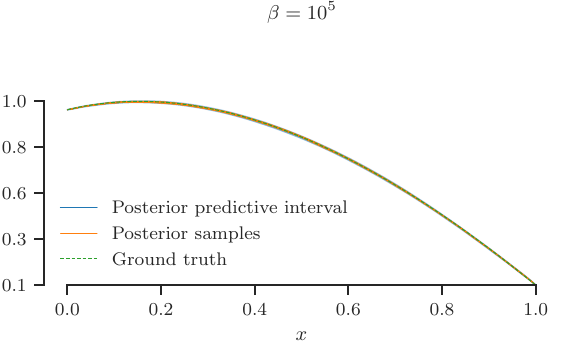}
        \label{fig:ex1d}
    \end{subfigure}
    \caption{Example 1 -- Effect of $\beta$ on the variance of the posterior.}
    \label{fig:ex1compiled}
\end{figure}

\subsection{Example 2 -- Inverse temperature is inversely proportional to model-form error}

In this example, we solve the inverse problem of identifying the inverse temperature $\beta$ for non-linear differential equations with controllable degree of model-form error.
We observe that PIFT yields $\beta$ posteriors that concentrate on large values when the model is correct and to low values when the model is wrong.

We generate synthetic data from the following nonlinear differential equation
\begin{equation}
    D\frac{d^2\phi}{dx^2}-\kappa\phi^3 = f,
    \label{eqn:nonlinear-pde}
\end{equation}
in $x$ in $[0,1]$, with boundary conditions:
$$
\phi(0) = 0,\;\text{and}\;\phi(1)=0.
$$
The parameters are set to $D = 0.1$, $\kappa=1$, and the ground truth source term to:
$$
f(x) = \cos(4x).
$$
Then, we take measurements of the exact solution at 40 equidistant points excluding the boundaries. The noise is Gaussian with variance $\sigma = 0.01^2$.

The energy for this problem is
\begin{equation}
    U[\phi] = \int_0^1dx \left[\frac{1}{2}D\left(\frac{d\phi}{dx}\right)^2 + \frac{1}{4}\kappa\phi^4+\phi f\right].
    \label{eqn:nonlinear-energy}
\end{equation}
Minimizing \qref{nonlinear-energy} constrained on any boundary conditions is  equivalent to solving \qref{nonlinear-pde}, see Appendix~\ref{appendix:proof_nonlinear_energy} for details.
We parameterize the field in exactly the same way as the previous example (boundary conditions are automatically satisfied, 20 terms in Fourier series expansion).

We design two separate experiments:

\paragraph{Experiment a (source term error):} We consider a source term of the form
$$
f(x;\gamma) = \gamma\cos(4x) + (1-\gamma)\exp(-x),
$$
and we vary $\gamma$ between $0$ and $1$.
When $\gamma=1$, PIFT uses the exact physics.
When $\gamma=0$, it uses the wrong physics.
Intermediate $\gamma$'s correspond to intermediate source term errors.

\paragraph{Experiment b (energy error):}
We consider a differential equation of the form
\begin{equation}
    D\frac{d^2\phi}{dx^2}-\gamma k\phi^3 -(1-\gamma)\phi = f,
    \label{eqn:nonlinear-pde-wrong}
\end{equation}
where again we vary $\gamma$  between 0 (completely incorrect physics) and 1 (correct physics).
We then train the model to learn the field governed by the ground truth physics \qref{nonlinear-pde} using the energy which comes from \qref{nonlinear-pde-wrong}, which is given by
$$
U[\phi;\gamma] = \int_0^1dx\left(\frac{1}{2}D\left|\frac{d\phi}{dx}\right|^2 + \gamma\frac{1}{4}\kappa\phi^4+\frac{1}{2}(1-\gamma)\phi^2+\phi f\right).
$$

\begin{figure}[htbp]
    \centering
    \begin{subfigure}[b]{0.475\textwidth}
        \centering
        \includegraphics[width=\textwidth]{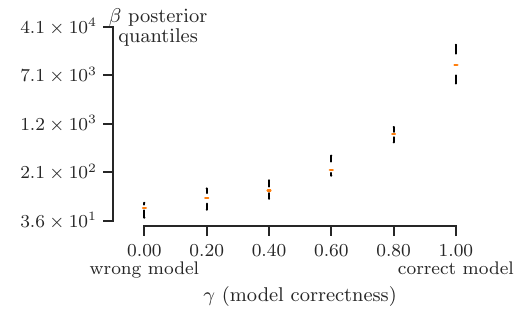}
        \caption{Experiment a (source term error).}
        \label{fig:ex2a}
    \end{subfigure}
    \hfill
    \begin{subfigure}[b]{0.475\textwidth}
        \centering
        \includegraphics[width=\textwidth]{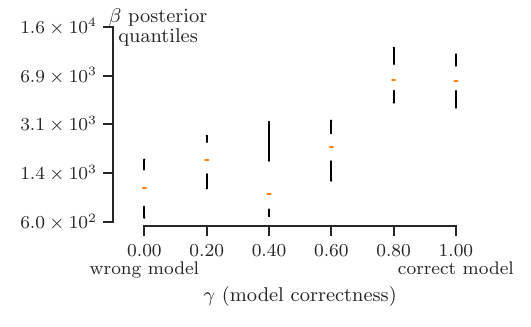}
        \caption{Experiment b (energy error).}
        \label{fig:ex2b}
    \end{subfigure}
    \caption{{Example 2 -- Sensitivity of $\beta$ on model correctness.}}
    \label{fig:ex2compiled}
\end{figure}

For each experiment, we take six equidistant points of $\gamma$ from 0 to 1. 
Each $\gamma$ specifies a model form.
For each model form, we use the available measurments to sample from the posterior of $\beta$ using SGLD.
Note that we use a Jeffrey's prior $p(\beta) \propto \frac{1}{\beta}$ and that we actually sample in log-space, i.e., the parameter we sample is $\lambda = \log\beta$.
The SGLD learning rate for $\lambda=\log\beta$ is set to $\alpha_0=10^{-3}$.
The learning rate for the posterior field sampling is $\hat{\alpha}_0=2$.
The rest of the parameters are as specified in~\sref{sec:sgld_params_1} and~\sref{sec:sgld_params_2}.

The results are shown in~\fref{ex2compiled}.
We observe a positive trend between $\beta$ and $\gamma$.
That is, we empirically verify that the model automatically places a greater trust in the physics when the known physics are closer to reality.
For smaller values of $\gamma$, the model recognizes that the physics are incorrect, and chooses to ignore them.
Then, the model automatically makes the decision to treat learning the field as a Bayesian regression problem, which is why we observe smaller values of $\beta$.
On the other end, when the model detects that physics are (mostly) correct, i.e. higher values of $\gamma$, it selects a larger $\beta$ so that it is able to learn from the physics.
We remark here that $\beta$ can be used as a metric to detect when the known physics need to be corrected, in which case we can attempt to derive a more accurate representation of the physics, such as relaxing any assumptions which may have been made when deriving equations that govern the field of interest.
Although these are only empirical studies, this approach contributes to the problem of quantifying \emph{model-form uncertainty}, an important emerging topic of research which is discussed in \cite{sargsyan2019embedded}.

\subsection{Example 3 -- Inverse problem involving non-linear differential equation}
\label{sec:inverseproblem}

In what follows the ground truth physics are the same as the previous example.
The differential equation is given in~\qref{nonlinear-pde}, the boundary conditions are $\phi(0)=\phi(1)=0$, the ground truth parameters are $D=0.1, \kappa=1$, and the ground truth source term is $f(x)=\cos(4x)$.
We pose and solve two separate inverse problems.
In~\sref{sec:example3a}, we seek to identify $D$ and $\kappa$.
In~\sref{sec:example3b}, we seek to identify $D$, $\kappa$, and $f$.

Several details are identical in both cases.
First, we use the same 40 observations as used in the previous example and we assume that we know the correct form of the field energy and the likelihood.
We set $\beta=10^5$ to indicate our strong trust on the model.
Second, we set Jeffrey's priors for $D$ and $\kappa$:
$$
p(D) \propto \frac{1}{D}\;\text{and}\;p(\kappa) \propto \frac{1}{\kappa}.
$$
Since these are positive parameters, we sample in the log-space, i.e.,
$$
\lambda_1 = \log (D)\;\text{and}\;\lambda_2 = \log(\kappa).
$$
The SGLD learning rate for sampling the parameters $\lambda$ is $\alpha_0 = 0.1$.
The SGLD learning rate for posterior field sampling is $\hat{\alpha}_0 = 10$.
The rest of the parameters are as in~\sref{sec:sgld_params_1} and~\sref{sec:sgld_params_2}.
Finally, we parameterize the field in exactly the same way as in the previous two examples, i.e.,
it automatically satisfies the boundary conditions and it uses a truncated Fourier series with 20 terms.

\subsubsection{Example 3a -- Identification of the energy parameters}\label{sec:example3a}

In this problem we seek to identify $D$ and $\kappa$ from the noisy measurements. The results are summarized in \fref{3acompiled}.
Subfigure~(a) shows every 10,000th sample of the SGLD iterations.
Notice that the SGLD Markov chain is mixing very slowly, albeit the computational cost per iteration is very small (100,000 iterations take about 1 second on a 2021 Macbook Pro with an Apple M1 Max chip and 64 GB of memory).
Subfigure~(b) shows the joint posterior over $D$ and $\kappa$.
We see that $D$ is identified to very good precision, but $\kappa$ is less identifiable.
Repeating the simulation several times, we observed that $D$ always gets a marginal posterior as in Subfigure~(b), but the marginal posterior of $\kappa$ may shift up or down depending on the starting point of SGLD.
This suggests that SGLD may not be mixing fast enough and this may be problematic in cases where the posterior does not have a single pronounced mode.
In other words, either more iterations are needed to fully explore the posterior along $\kappa$ or a different learning rate is required in the $\kappa$ direction.
In Subfigure~(c) we show what we call the \emph{fitted prior predictive} distribution.
This distribution shows what PIFT thinks the prior is after it has calibrated the parameters.
Mathematically, it is:
$$    
p_{\text{fitted prior pred.}}(\phi|d) = \int d\lambda\:p(\phi|\lambda)p(\lambda|d).
$$
The fitted prior predictive is what PIFT tells us what our beliefs about the field should be in a situation with identical boundary conditions but before we observe any data.
Subfigure~(d) depicts the posterior predictive which is:
$$    p(\phi|d) = \int d\lambda\: p(\phi|d,\lambda)p(\lambda|d).
$$
This is what PIFT says the field is when we consider all the data for this particular situation.
Intuitively, PIFT attempts to fit the physical parameters so that the data are mostly explained by the fitted prior predictive, i.e., from the physics.
This because we put a strong a priori belief in the physics.

\begin{figure}[htbp]
    \centering
    \begin{subfigure}[b]{0.475\textwidth}
        \centering
        \includegraphics[width=\textwidth]{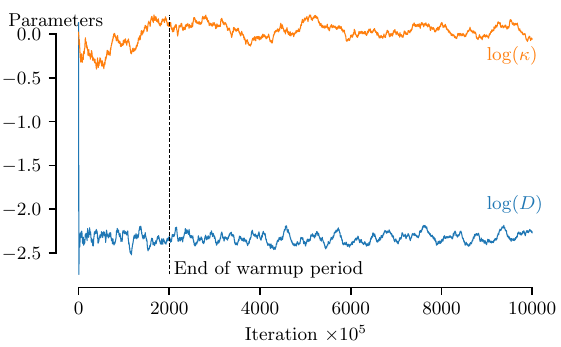}
        \caption{Trace of SGLD Markov chain.}
        \label{fig:ex3a_trace}
    \end{subfigure}
    \hfill
    \begin{subfigure}[b]{0.475\textwidth}
        \centering
        \includegraphics[width=\textwidth]{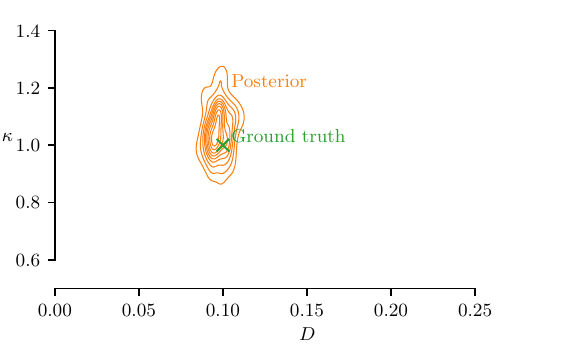}
        \caption{Joint posterior of $D$ and $\kappa$.}
        \label{fig:ex3a_post_params}
    \end{subfigure}
    \vskip\baselineskip
    \begin{subfigure}[b]{0.475\textwidth}
        \centering
        \includegraphics[width=\textwidth]{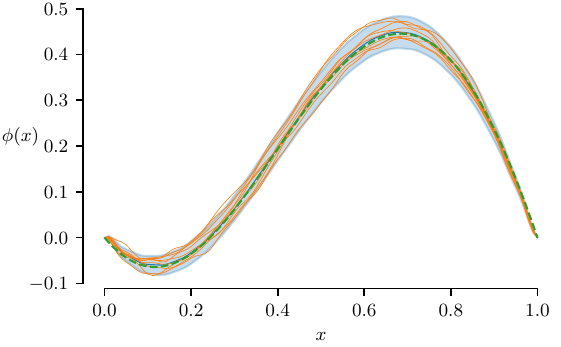}
        \caption{Fitted prior predictive.}
        \label{fig:ex3a_prior}
    \end{subfigure}
   \hfill
    \begin{subfigure}[b]{0.475\textwidth}
        \centering
        \includegraphics[width=\textwidth]{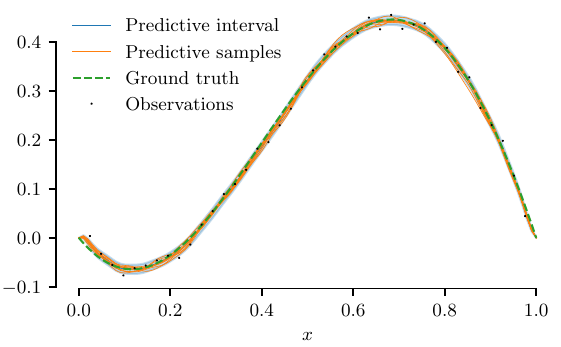}
        \caption{Posterior predictive.}
        \label{fig:ex3a_field_post}
    \end{subfigure}
    \caption{Example 3a -- Identification of energy parameters.}
    \label{fig:3acompiled}
\end{figure}

\subsubsection{Example 3b -- Simultaneous identification of the energy parameters and the source term}\label{sec:example3b}

Next, we solve the same problem as before, but in this case we seek to infer the source term along with the field, $D$, and $\kappa$.
We assume the source is a priori a zero mean, Gaussian random field with covariance $C$,
$$
p(f) = \mathcal{N}(f|0, C).
$$
To encode our beliefs about the variance (equal to one), smoothness (infinitely differentiable), and lengthscale (equal to $0.3$), we take $C$ to be the squared exponential:
$$
C(x,x') = \exp\left\{-\frac{(x-x')^2}{2(0.3)^2}\right\}.
$$
Because we need to work with a finite number of parameters, we take the Karhunen-Lo\`eve expansion (KLE) of $f$ and truncate it at 10 terms.
This number of terms is sufficient to explain more than 99\% of the energy of the field $f$.
We construct KLE numerically using the Nystr\"om approximation as explained in~\cite{bilionis2016bayesian}.
So, the parameters $\lambda$ we seek to identify include $\lambda(D),\lambda(\kappa)$, and the 10 KLE source terms.

The results are summarized in \fref{3bcompiled}.
As in the previous example, we observe that $D$ is more identifiable than $\kappa$, see Subfigure~(a).
Repeated runs reveal that the marginal posterior over $\kappa$ may shift up and down indicating slow SGLD mixing.
The source term, see Subfigure~(b), is identified relatively well, albeit this is probably due to the very informative prior which penalizes small lengthscales.
When we repeat the same experiment under the assumption that the prior of the source term has a smaller lengthscale, we observe a bigger spread in the joint $D$-$\kappa$ posterior and a discrepancy of the source term at the boundaries.
This is despite the fact that the fitted prior predictive, see Subfigure~(c), and the posterior predictive, see Subfigure~(d), are always identical to the results presented here.
This is evidence that the $D$, $\kappa$ and $f$ are not uniquely identifiable from the observed data without some prior knowledge.

\begin{figure}[htbp]
    \centering
    \begin{subfigure}[b]{0.475\textwidth}
        \centering
        \includegraphics[width=\textwidth]{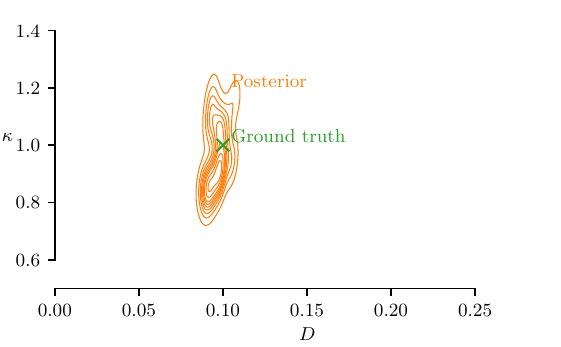}
        \caption{Joint posterior over $D$ and $\kappa$.}
        \label{fig:ex3b_post_params}
    \end{subfigure}
    \hfill
    \begin{subfigure}[b]{0.475\textwidth}
        \centering
        \includegraphics[width=\textwidth]{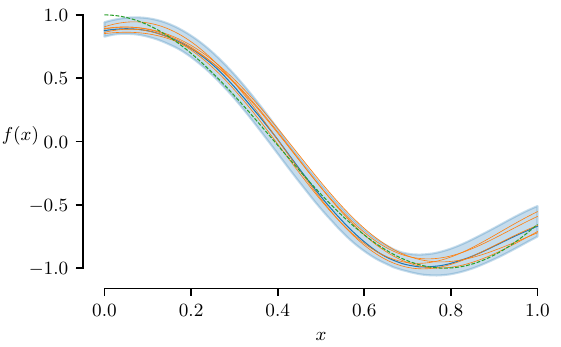}
        \caption{Posterior of source term}
        \label{fig:ex3b_source}
    \end{subfigure}
    \vskip\baselineskip
    \begin{subfigure}[b]{0.475\textwidth}
        \centering
        \includegraphics[width=\textwidth]{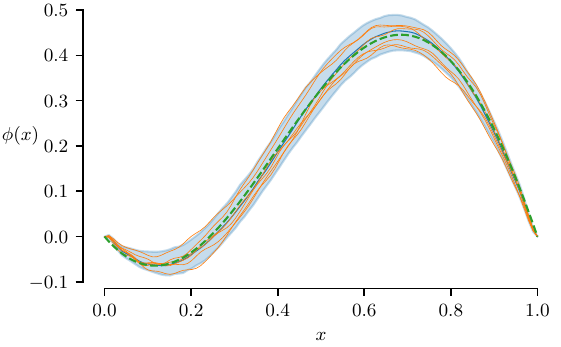}
        \caption{Fitted prior predictive.}
        \label{fig:ex3b_prior}
    \end{subfigure}
   \hfill
    \begin{subfigure}[b]{0.475\textwidth}
        \centering
        \includegraphics[width=\textwidth]{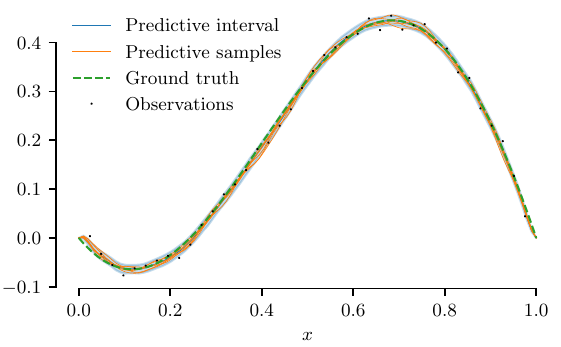}
        \caption{Posterior predictive.}
        \label{fig:ex3b_field_post}
    \end{subfigure}
    \caption{Example 3b -- Simultaneous identification of energy parameters and source term.}
    \label{fig:3bcompiled}
\end{figure}

Remember that in this example we have fixed $\beta$ to $10^5$.
However, in this simultaneous parameter and source identification experiment, it is possible to extract a common factor out of all the field energy terms.
This means that we are actually allowing PIFT to change the effective $\beta$ from the nominal value.
For this particular choice of source term prior, the problem seems well-posed and $\beta=10^5$ is very likely close to the posterior mode of $\beta$.
However, in our numerical simulations (not reported in this paper) with less informative source term priors (e.g., smaller lengthscale and more KLE terms), we observed that there may be many combinations of $D$, $\kappa$, and $f$ that yield the same fitted prior predictive and posterior predictive and the problem becomes ill-posed.
It is not possible to remove this indeterminacy without adding some observations of the source term.

\subsection{Example 4 -- Ill-posed 2D forward problems}

We provide an example of an ill-posed forward problem, and show how the field posterior captures the possible solutions.
We study the 2D time-independent Allen-Cahn equation, given by
\begin{equation}
    \varepsilon\nabla^2\phi -  \phi(\phi^2-1) + f = 0
    \label{eqn:ACpde}
\end{equation}
on the domain $\Omega = [-1,1]\times[-1,1]$, where $\varepsilon$ is a parameter known as the mobility, and $f$ is the source term.
The Allen-Cahn equation is a classic PDE which models certain reaction-diffusion systems involving phase separation.
We select $\varepsilon=0.01$, and employ the ground truth solution
$$
\phi(x,y) = 2\exp\left\{-(x^2+y^2)\right\}\sin(\pi x)\sin(\pi y),
$$
which then specifies the source term and boundary conditions.

One can show that the energy functional associated  with \qref{ACpde} is
\begin{equation}
    U_{\varepsilon}[\phi] = \int_{\Omega}d\Omega\left(\frac{\varepsilon}{2}||\nabla\phi||^2 + \frac{1}{4}(1-\phi^2)^2 - f\phi\right),
    \label{eqn:ACenergy}
\end{equation}
where the critical points of \qref{ACenergy} are solutions to \qref{ACpde} \cite{pacard2009geometric}.
\qref{ACenergy} has multiple critical points (in some cases saddle points, depending on the parameters), which has the possibility to define an ill-posed problem as the solutions to \qref{ACpde} may not be unique.
To demonstrate this, we parameterize the field with a ``well-informed'' basis with a single parameter given by
$$
\hat{\phi}(x,y;\theta) = \theta\exp\left\{-(x^2+y^2)\right\}\sin(\pi x)\sin(\pi y),
$$
where we know $\theta$ should be exactly $2$ to match the ground truth.
Taking this parameterization and inserting it into \qref{ACenergy}, we derive the physics-informed prior given by
\begin{equation}
    p(\theta)\propto \exp\left\{-\beta U_{0.01}(\theta)\right\}.
    \label{eqn:ACsimpleprior}
\end{equation}
Because there is only a single parameter, we can visualize the prior, as seen in \fref{ACprior}.
\begin{figure}[htbp]
  \centering
  \includegraphics{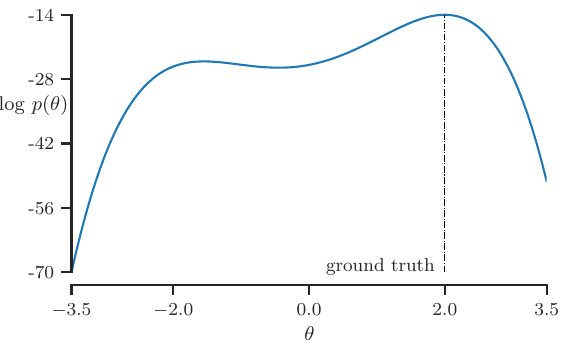}
  \caption{Example 4 -- Physics-informed functional prior \qref{ACsimpleprior} with $\beta=100$.}
  \label{fig:ACprior}
\end{figure}
We observe that the physics-informed prior coming from the Allen-Cahn energy is bimodal.
Given insufficient data, the posterior will remain bimodal.
This results in the possibility to easily produce an ill-posed problem, where the classic PDE solvers struggle to capture the true solution.
We demonstrate that PIFT is equipped to handle this scenario. Even if the problem is not well-posed the ground truth solution can be captured by one of the posterior modes.

To pose the remainder of the problem, we combine \qref{ACpde} with noisy measurements of the exact solution on only $3$ boundaries.
On each of these boundaries, we sample $15$ uniformly-distributed locations and take noisy measurements of the field.
The noise is Gaussian with variance $\sigma^2=0.01^2$.
This results in an ill-posed problem, as we are missing measurements on one of the boundaries.

We parameterize the field with the 2D Fourier series defined on $[-1,1]\times[-1,1]$, which we denote by $\hat{\phi}(x,y;\theta)$.
For details on the definition of the real-valued Fourier series in multiple dimensions, see \cite{davis1986statistics}.
In this problem, we find that $9$ total Fourier terms to define $\hat{\phi}$ is sufficient to capture both posterior modes without creating issues in sampling.
In theory, PIFT works when we use more Fourier terms, albeit in practice the Markov chain may have to take an impractical number of steps to jump between posterior modes if the energy barrier between them is very high.

By inserting the 2D Fourier series, $\hat{\phi}$, into the Allen-Cahn energy defined in \qref{ACenergy}, we obtain the physics-informed prior over the Fourier coefficients, $\theta$.
The likelihood is obtained through evaluating $\hat{\phi}$ at the locations of the measurements in $\Omega$, and the posterior over $\theta$ is obtained via application of Bayes's rule.

To sample from the posterior, we employ the HMCECS sampling scheme as outlined in \sref{sec:numericalposts}.
Each sample generated from the posterior requires evaluation of the energy, which involves an integral in 2D space.
To approximate this integral numerically, we implement a minibatch subsampling scheme.
Specifically, at each step we randomly select a minibatch of points in $(x_i,y_i)\in\Omega$, $i=1,\dots,b$, where the points $(x_i,y_i)$ follow a uniform distribution over the domain.
Given a sufficient number of samples, the posterior obtained through the subsampling scheme remains invariant to the posterior as if the integral was evaluated in the continuous sense due to the properties of the HMCECS algorithm.
This subsampling scheme provides the benefit of significantly improving the computational efficiency of the algorithm.

We select a minibatch size of $b=8$ and generate $100,000$ samples from the posterior through HMCECS with a No-U-Turn sampler selected as the inner kernel \cite{hoffman2014no}.
As we are solving a forward problem, we fix $\beta$ to $50$.
This value is high enough to enforce the physics, but not so high that the non-dominating modes of the field posterior disappear.
In \fref{ACposts}, we show the posteriors over a small selection of the Fourier coefficients.
\begin{figure}[htbp]
  \centering
  \begin{subfigure}[t]{0.45\textwidth}
  \centering
  \includegraphics[width=\textwidth]{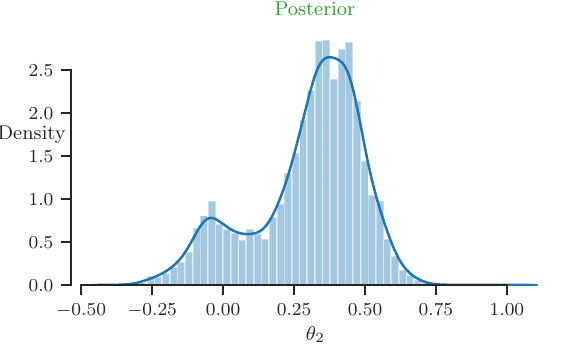}
  \label{fig:theta2samps}
  \end{subfigure} \hfill
  \begin{subfigure}[t]{0.45\textwidth}
  \centering
  \includegraphics[width=\textwidth]{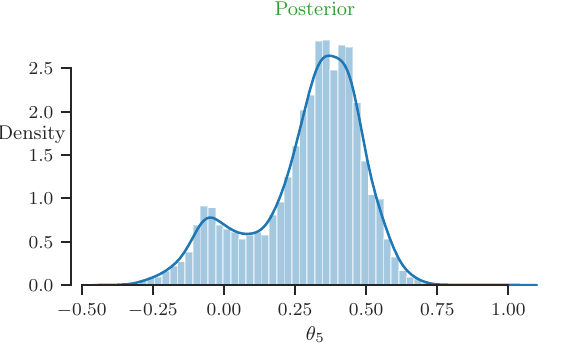}
  \label{fig:theta5samps}
  \end{subfigure}
  \caption{Example 4 -- Selected field parameter marginal posteriors exhibiting bimodality.}
  \label{fig:ACposts}
\end{figure}
Here, we clearly observe a bimodal form of the posterior, which arises from the fact that the problem is not well-posed.
Because the posterior contains two modes, naively selecting the mean or median to approximate the solution will result in a model which fails.

To make predictions, we separate the modes of the posterior by employing a Gaussian mixtures (GM) model, which approximates the posterior as a combination of Gaussian distributions \cite{reynolds2009gaussian}.
This GM model allows us to make separate predictions from each of the modes.
\begin{figure}[htbp]
    \centering
    \begin{subfigure}[b]{0.475\textwidth}
        \centering
        \includegraphics[width=\textwidth]{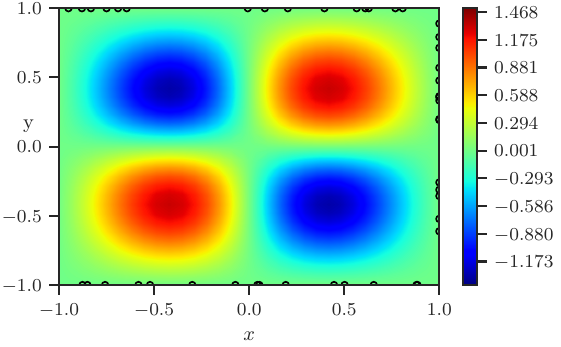}
        \caption{ground truth}
        \label{fig:ex4a}
    \end{subfigure}
    \hfill
    \begin{subfigure}[b]{0.475\textwidth}
        \centering
        \includegraphics[width=\textwidth]{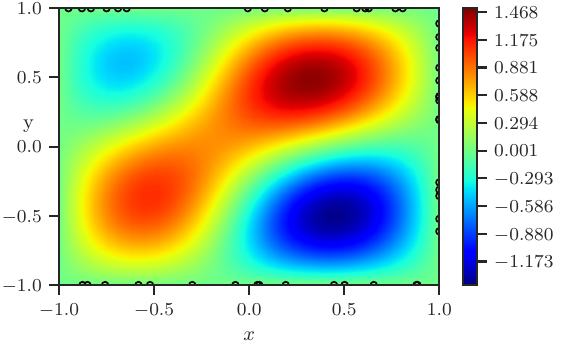}
        \caption{median sample}
        \label{fig:ex4b}
    \end{subfigure}
    \vskip\baselineskip
    \begin{subfigure}[b]{0.475\textwidth}
        \centering
        \includegraphics[width=\textwidth]{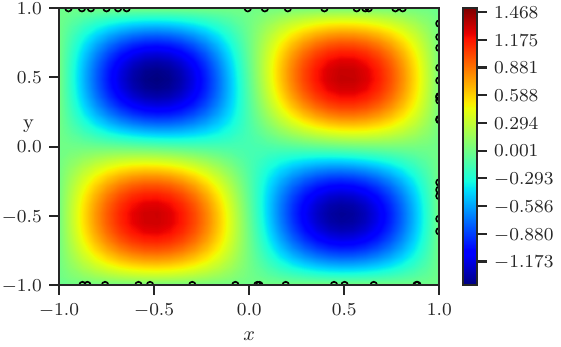}
        \caption{1st mode}
        \label{fig:ex4c}
    \end{subfigure}
   \hfill
    \begin{subfigure}[b]{0.475\textwidth}
        \centering
        \includegraphics[width=\textwidth]{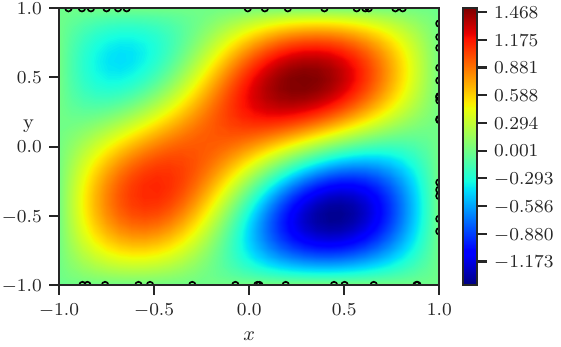}
        \caption{2nd mode}
        \label{fig:ex4d}
    \end{subfigure}
    \caption{Example 4 -- Predictions from each of the posterior modes. Here, we find that the first mode solves the problem, while the second fails. We also see that naively selecting the median of the full posterior fails to capture the solution, as the posterior is bimodal. The locations of the measurements are marked on the boundaries.}
    \label{fig:ACsolns}
\end{figure}
In \fref{ACsolns}, we clearly see that one mode provides much better predictions, while the other mode (and likewise the median) fails.
The absolute errors for each mode, the median and mean are shown in \fref{ACerrors}.
\begin{figure}[htbp]
    \centering
    \begin{subfigure}[b]{0.475\textwidth}
        \centering
        \includegraphics[width=\textwidth]{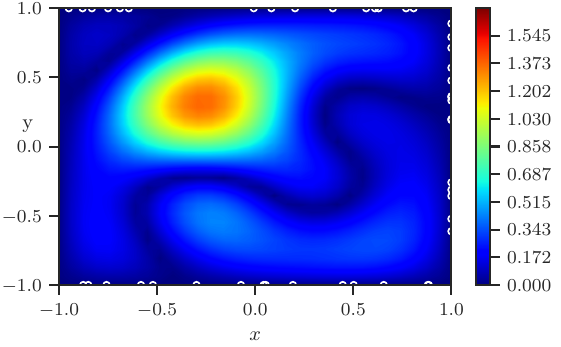}
        \caption{mean}
        \label{fig:ex4e}
    \end{subfigure}
    \hfill
    \begin{subfigure}[b]{0.475\textwidth}
        \centering
        \includegraphics[width=\textwidth]{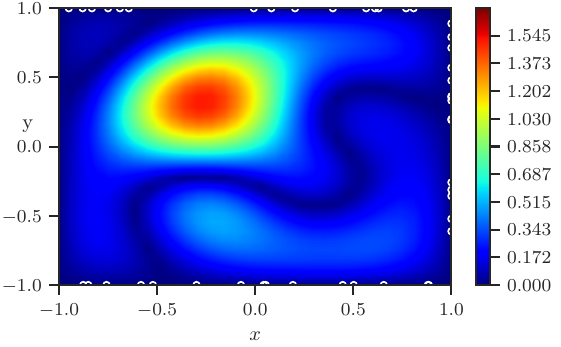}
        \caption{median}
        \label{fig:ex4f}
    \end{subfigure}
    \vskip\baselineskip
    \begin{subfigure}[b]{0.475\textwidth}
        \centering
        \includegraphics[width=\textwidth]{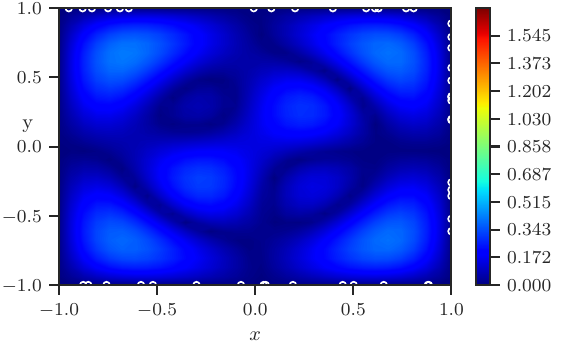}
        \caption{1st mode}
        \label{fig:ex4g}
    \end{subfigure}
   \hfill
    \begin{subfigure}[b]{0.475\textwidth}
        \centering
        \includegraphics[width=\textwidth]{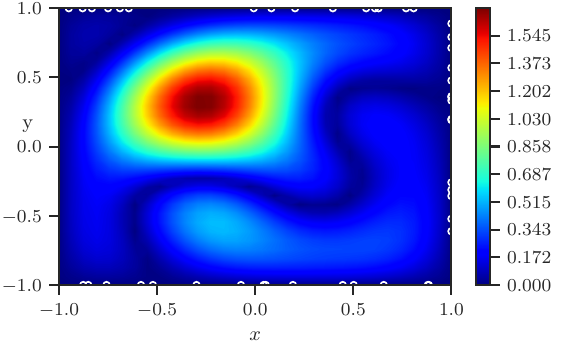}
        \caption{2nd mode}
        \label{fig:ex4h}
    \end{subfigure}
    \caption{Example 4 -- Absolute errors for the mean, median, and modes of the posterior.}
    \label{fig:ACerrors}
\end{figure}
We find that the error for the first mode remains reasonable, given the fact that the problem is ill-defined.
This is a remarkable result, as the PIFT approach is able to accurately capture the solutions to ill-posed problems in an effortless manner by separating the different modes of the posterior.

\section{Conclusions}
\label{sec:conclusions}

Using physics-informed information field theory, we have developed a method to perform Bayesian inference over physical systems in a way which combines data with known physical laws.
By encoding knowledge of the physics into a physics-informed functional prior, we are able to derive a posterior over the field of interest in a way which seamlessly combines measurements of the field with the known physics.
The probability measures presented here are defined over function spaces, and they remain independent of any field discretization.
We also find that under certain conditions, analytic representations of the functional priors and field posteriors can be derived, where the mean and covariance operators are controlled by the Green's function of the corresponding partial differential equation.
To numerically approximate the posteriors, we developed a nested stochastic gradient Langevin dynamics approach. At the outer layer one samples the energy parameters while in the inner layers one samples from the prior and posterior of the fields.
Further, we highlighted how PINNs and B-PINNs for reconstructing fields and solving inverse problems are related to PIFT.
We find that, under certain conditions, PINNs is a maximum a posteriori estimate of PIFT.
B-PINNs differs fundamentally from PIFT in that it enforces the physics through observations instead of the prior.
Using results from Bayesian asymptotic analysis we showed that B-PINNs yields field posteriors that collapse to a single solution as the number of collocation points goes to infinity.
In contrast, PIFT's posterior uncertainty is controlled by the inverse temperature parameter $\beta$, a fact that we demonstrated both through an analytical example and in numerical examples.
In our numerical examples, we also highlighted the ability of PIFT to solve Bayesian inverse problems.
Finally, we demonstrated that PIFT posteriors may not necessarily remain unimodal, as they can capture multiple possible solutions for ill-posed problems.

There remain a number of open problems regarding the method.
First, analytic representations of the functional priors and posteriors were able to be derived for quadratic operators.
Ideally, this could be generalized to other operators.
Doing so requires perturbation approximations of the path integrals that appear or through the rich theory of Feynman diagrams \cite{feynman2010quantum}.
The application of Feynman diagrams to classical IFT has been explored in \cite{pandey2022review} for Gaussian random fields.

Because the model is able to detect when the representation of the physics is incorrect, this contributes to the problem of quantifying model-form uncertainty.
However, the exact nature of this contribution is not completely understood, and further analytical and empirical studies could prove to be insightful.
The theory presented here is developed only for deterministic problems without dynamics, and it would be beneficial to study time-dependent problems with stochastic state transitions.
Because the overall goal of this paper is to introduce the theory, problems coming from real physical systems were not presented, and this leaves gaps for real-world applications to be explored.
As Fourier series in 1D or 2D was the primary choice of surrogate functions in this paper, it may be useful to study how other choices perform.
There are certain applications for which the Fourier series is a bad choice.
For example, in situations for which the field has discontinuities or ``jumps'', the Fourier series will fail to accurately capture the discontinuities due to the notorious and well-documented Gibbs phenomenon.
In this type of situation a different surrogate which can capture discontinuities must be used, such as a deep neural network with an appropriate activation function.
Finally, the approach used here to numerically approximate the field posteriors was primarily stochastic gradient Langevin dynamics, and it may be interesting to develop different approaches based on other schemes such as a variational inference approach \cite{blei2017variational}.

\appendix
\section{Proofs of \qref{expect_of_field} and \qref{cov_of_field}}
\label{appendix:expectation_and_cov_prior}
We begin by showing that \qref{field_prior} defines a zero-mean field by evaluating the expectation given by \qref{expect_of_field}.
To make progress, we show that this expression relates to the functional derivative of the partition function with respect to $q$.
We define $\frac{\delta Z[q]}{\delta q(x)}$ to be the functional derivative of $Z[q]$ with respect to $q$ in the direction of the Dirac delta function $\delta_x(x') = \delta(x'-x)$, i.e.,
$$
\frac{\delta Z[q]}{\delta q(x)} = \lim_{\epsilon\rightarrow 0}\frac{Z[q+\epsilon\delta_x] - Z[q]}{\epsilon}.
$$
We have:
$$
\begin{aligned}
\frac{\delta Z[q]}{\delta q(x)} &=
\frac{\delta }{\delta q(x)} \int \calD\phi\: \exp\left\{-\frac{1}{2}\phi^\dagger S\phi + \phi^\dagger q\right\}\\
&= 
\int \calD\phi\:
\frac{\delta }{\delta q(x)}\exp\left\{-\frac{1}{2}\phi^\dagger S\phi + \phi^\dagger q\right\}\\
&=
\int \calD\phi\:
\exp\left\{-\frac{1}{2}\phi^\dagger S\phi + \phi^\dagger q\right\}
\frac{\delta }{\delta q(x)}\left[\phi^\dagger q\right]\\
&=
\int \calD\phi\:
\exp\left\{-\frac{1}{2}\phi^\dagger S\phi + \phi^\dagger q\right\}
\frac{\delta }{\delta q(x)}\left[\int dx' \phi(x') q(x')\right]\\
&=
\int \calD\phi\:
\exp\left\{-\frac{1}{2}\phi^\dagger S\phi + \phi^\dagger q\right\}
\left[\int dx' \phi(x') \frac{\delta }{\delta q(x)}q(x')\right]\\
&= 
\int \calD\phi\:
\exp\left\{-\frac{1}{2}\phi^\dagger S\phi + \phi^\dagger q\right\}
\left[\int dx' \phi(x') \delta(x-x')\right]\\
&= \int \calD\phi\:
\exp\left\{-\frac{1}{2}\phi^\dagger S\phi + \phi^\dagger q\right\}
\phi(x).
\end{aligned}
$$
The right hand side divided by the normalization constant $Z[q=0]$ is the expectation of the field.
So, we have shown that:
\begin{equation}
\label{eqn:intermediate}
    \mathbb{E}[\phi(x)] = Z[q=0]^{-1}\frac{\delta Z[q=0]}{\delta q(x)}.
\end{equation}
Since we already have an analytical expression for Z[q], \qref{partition_analytical}, we can also evaluate the right hand side.
To this end:
$$
\begin{aligned}
\frac{\delta Z[q]}{\delta q(x)} &= 
\frac{\delta}{\delta q(x)} C\left[\det S\right]^{-1/2}\exp\left\{\frac{1}{2} q^\dagger S^{-1}q\right\}\\
&=
C\left[\det S\right]^{-1/2}\exp\left\{\frac{1}{2} q^\dagger S^{-1}q\right\}
\frac{\delta }{\delta q(x)}
\frac{1}{2}q^\dagger S^{-1}q\\
&=
\frac{1}{2}Z[q]
\frac{\delta }{\delta q(x)}
\int dx' dx''\: q(x') S^{-1}(x', x'')q(x'')\\
&=
\frac{1}{2}Z[q]
\int dx' dx''\: \left[\delta(x'-x) S^{-1}(x', x'')q(x'')
+ q(x')S^{-1}(x', x'')\delta(x''-x)
\right]\\
&=
\frac{1}{2}Z[q]\left[\int dx''\: S^{-1}(x, x'')q(x'') + \int dx'\:q(x')S^{-1}(x', x)\right]\\
&= Z[q] (S^{-1}q)(x).
\end{aligned}
$$
Plugging in \qref{intermediate}, we get:
$$
\mathbb{E}[\phi(x)] = (S^{-1}q)(x)|_{q=0} = 0,
$$
which completes the proof.

Next, we show that the covariance of the field is simply the inverse of the operator $S$.
Start by noticing that:
$$
\begin{aligned}
    \frac{\delta^2 Z[q]}{\delta q(x)\delta q(x')}
    &= \frac{\delta }{\delta q(x')}\frac{\delta Z[q]}{\delta q(x)}\\
&= \frac{\delta }{\delta q(x')}\int \calD\phi\:
\exp\left\{-\frac{1}{2}\phi^\dagger S\phi + \phi^\dagger q\right\}
\phi(x)\\
&=
\int\calD\phi\:
\exp\left\{-\frac{1}{2}\phi^\dagger S\phi + \phi^\dagger q\right\}
\phi(x) \frac{\delta }{\delta q(x')}[\phi^\dagger q]\\
&=
\int\calD\phi\:
\exp\left\{-\frac{1}{2}\phi^\dagger S\phi + \phi^\dagger q\right\}
\phi(x) \phi(x').
\end{aligned}
$$
If we divide the right-hand side by the normalization constant $Z[q=0]$ we get:
\begin{equation}
    \label{eqn:covariance_tmp}
    \mathbb{E}[\phi(x)\phi(x')] = Z[q=0]^{-1}\frac{\delta^2 Z[q=0]}{\delta q(x)\delta q(x')}.
\end{equation}
To evaluate the right hand side we need:
$$
\begin{aligned}
    \frac{\delta^2 Z[q]}{\delta q(x)\delta q(x')}
    &= \frac{\delta }{\delta q(x')}\left\{Z[q] (S^{-1}q)(x)\right\}\\
    &=
    \frac{\delta Z[q]}{\delta q(x')}(S^{-1}q)(x) + Z[q]\frac{\delta}{\delta q(x')}(S^{-1}q)(x)\\
    &=
    Z[q](S^{-1}q)(x')(S^{-1}q)(x) + Z[q] \frac{\delta}{\delta q(x')}\int dx''\: S^{-1}(x,x'')q(x'')\\
    &=
        Z[q](S^{-1}q)(x')(S^{-1}q)(x) + Z[q] \int dx''\: S^{-1}(x,x'')\delta(x''-x')\\
    &=
        Z[q](S^{-1}q)(x')(S^{-1}q)(x) + Z[q] S^{-1}(x,x').
\end{aligned}
$$
Dividing by the normalization constant $Z[q=0]$, evaluating at $q=0$, and using \qref{covariance_tmp}, we get:
$$
    \mathbb{E}[\phi(x)\phi(x')] = S^{-1}(x,x'),
$$
which proves our assertion.

\section{Proof of \qref{unbiased_grad_lambda}}
\label{appendix:unbiased_grad_lambda_derivation}

We have:
$$
\frac{\partial H(\lambda|d)}{\partial\lambda_i} =  -\frac{\partial }{\partial \lambda_i}\log p(\lambda|d) = -\frac{\partial }{\partial \lambda_i} \log\int_{\Phi}\mathcal{D}\phi\: p(\phi,\lambda|d).
$$
Using the chain rule, gives:
\begin{equation}
\label{eqn:step1}
\frac{\partial H(\lambda|d)}{\partial\lambda_i} = -\frac{\frac{\partial }{\partial \lambda_i}\int_{\Phi}\mathcal{D}\phi\;p(\phi,\lambda|d)}{\int_{\Phi}\mathcal{D}\phi\;p(\phi,\lambda|d)}
=-\frac{\frac{\partial }{\partial \lambda_i}\int_{\Phi}\mathcal{D}\phi\;p(\phi,\lambda|d)}{p(\lambda|d)}.
\end{equation}
The numerator is:
$$
\frac{\partial }{\partial \lambda_i}\int_{\Phi}\mathcal{D}\phi\;p(\phi,\lambda|d) = \frac{\partial }{\partial \lambda_i}\int_{\Phi}\mathcal{D}\phi\: \frac{1}{p(d)}p(d|\phi)\frac{\exp\left\{-\left[H[\phi|\lambda]+H(\lambda)\right]\right\}}{Z(\lambda)}.
$$
Passing the differentiation operator inside the integral, using the product rule and the chain rule, yields:
\begin{align*}
\frac{\partial }{\partial \lambda_i}\int_{\Phi}\mathcal{D}\phi\;p(\phi,\lambda|d) = 
\int_{\Phi}\mathcal{D}\phi\:\frac{1}{p(d)}p(d|\phi)\Bigg\{
&-\frac{\exp\left\{-\left[H[\phi|\lambda]+H(\lambda)\right]\right\}}{Z(\lambda)}\left[\frac{\partial H[\phi|\lambda]}{\partial\lambda_i}+\frac{\partial H(\lambda)}{\partial\lambda_i}\right]\\
&- \frac{\exp\left\{-\left[H[\phi|\lambda]+H(\lambda)\right]\right\}}{Z(\lambda)}\frac{1}{Z(\lambda)}\frac{\partial Z(\lambda)}{\partial\lambda_i}
\Bigg\}.
\end{align*}
Now, notice the following.
The ratio that appears in both terms, multiplied by $\frac{1}{p(d)}p(d|\phi)$, is the joint of $p(\phi,\lambda|d)$.
If we divide the joint by the denominator of \qref{step1}, we get the posterior of $\phi$ conditional on $\lambda$:
$$
p(\phi|d,\lambda) = \frac{p(\phi,\lambda|d)}{p(\lambda|d)}.
$$
Thus, we have shown that:
$$
\frac{\partial }{\partial \lambda_i}\int_{\Phi}\mathcal{D}\phi\;p(\phi,\lambda|d) = -\mathbb{E}\left[\frac{\partial H[\phi|\lambda]}{\partial\lambda_i}+\frac{\partial H(\lambda)}{\partial \lambda_i} + \frac{1}{Z(\lambda)}\frac{\partial Z(\lambda)}{\partial\lambda_i}\middle|d,\lambda\right].
$$
The two terms that do not depend on $\phi$ can be taken out of the expectation:
$$
\frac{\partial }{\partial \lambda_i}\int_{\Phi}\mathcal{D}\phi\;p(\phi,\lambda|d) = -\mathbb{E}\left[\frac{\partial H[\phi|\lambda]}{\partial\lambda_i}\middle|d,\lambda\right]
-\frac{\partial H(\lambda)}{\partial \lambda_i} - \frac{1}{Z(\lambda)}\frac{\partial Z(\lambda)}{\partial\lambda_i}.
$$
Finally, by direct differentiation of the partition function we can show that:
\begin{equation}
\label{eqn:deriv_Z}
\frac{1}{Z(\lambda)}\frac{\partial Z(\lambda)}{\partial\lambda_i} = -\frac{1}{Z(\lambda)}\int_{\Phi}\mathcal{D}\phi\: \exp\left\{-H[\phi|\lambda]\right\}\frac{\partial H[\phi|\lambda]}{\partial\lambda_i} = -\int_{\Phi}\mathcal{D}\phi\: p(\phi|\lambda)\frac{\partial H[\phi|\lambda]}{\partial\lambda_i},
\end{equation}
which completes the proof.

\section{Proof of \qref{nonlinear-energy}}
\label{appendix:proof_nonlinear_energy}
In order to show that \qref{nonlinear-energy} is the correct energy functional for \qref{nonlinear-pde}, we must show that functions which make this energy stationary are solutions to the differential equation.
We will proceed in two steps: first we will show that a function $\phi^*$ which solves \qref{nonlinear-pde} causes the first variation of \qref{nonlinear-energy} to vanish, i.e., $\phi^*$ is a critical point.
Then, we will show that the second variation of \qref{nonlinear-energy} is strictly positive-definite at $\phi^*$, i.e., $\phi^*$ is a local minimum of the energy.

To begin, we take the first variation of \qref{nonlinear-energy}.
By definition, this is
$$
\frac{\delta U[\phi]}{\delta \eta} = \frac{d}{d\epsilon}U[\phi+\epsilon\eta]|_{\epsilon=0},
$$
for an arbitrary function $\eta$ which is zero at the boundary.
To find the critical points of $U[\phi]$ we simply take $\delta U[\phi,\eta]=0$.
The first variation is:
\begin{equation}
    0 = \frac{\delta U[\phi]}{\delta \eta} = \int_0^1dx\:\left(D\frac{d\phi}{dx}\frac{d\eta}{dx}+\kappa\phi^3\eta+\eta f\right).
    \label{eqn:intermediate_var}
\end{equation}
Notice that through integration by parts
$$
\int_0^1dx\:\frac{d\phi}{dx}\frac{d\eta}{dx}=\frac{d}{dx}\left(\frac{d\phi}{dx}\eta\right)\Big|_0^1-\int_0^1 dx\:\frac{d^2\phi}{dx^2}\eta=-\int_0^1 dx\:\frac{d^2\phi}{dx^2}\eta
$$
since $\eta$ is zero on the boundaries.
Then \qref{intermediate_var} is equivalent to 
\begin{equation}
    0 = \int_0^1dx\:\left(-D\frac{d^2\phi}{dx^2}+\kappa\phi^3+f\right)\eta,
    \label{eqn:firstvariation}
\end{equation}
and in order for \qref{firstvariation} to hold for an arbitrary $\eta$ we must have that
$$
D\frac{d^2\phi}{dx^2}-\kappa\phi^3=f,
$$
which is exactly the differential equation given by \qref{nonlinear-pde}.
Therefore, the solution $\phi^*$ to \qref{nonlinear-pde} is a critical point of the functional given by \qref{nonlinear-energy}.

Next, we study the second variation of $U[\phi]$.
By definition this is
$$
\frac{\delta^2 U[\phi]}{\delta \eta^2}=
\frac{\delta}{\delta \eta}\left\{\frac{\delta U[\phi]}{\delta \eta}\right\}.
$$
We have
$$
\begin{aligned}
    \frac{\delta^2 U[\phi]}{\delta \eta^2} &= \frac{\delta}{\delta \eta}\left\{\int_0^1dx\:\left(D\frac{d\phi}{dx}\frac{d\eta}{dx}+\kappa\phi^3\eta+f\eta\right)\right\} \\
    &= \int_0^1dx\:\left(D\left(\frac{d\eta}{dx}\right)^2+3\kappa\phi^2\eta^2\right) \\
    &> \int_0^1dx\:D\left(\frac{d\eta}{dx}\right)^2 \\
    &> D\int_0^1dx\:\left(\frac{d\eta}{dx}\right)^2 \\
    &\geq Dc\int_0^1dx\:\eta^2 = Dc||\eta||^2,
\end{aligned}
$$
for some real number $c>0$, where the last inequality holds by the Poincar\'{e} inequality \cite{poincare1890equations}.
We have shown that the second variation of \qref{nonlinear-energy} is strictly positive for any $\phi$.
Then, the following two conditions hold
\begin{itemize}
    \item $\frac{\delta U[\phi^*]}{\delta \eta}=0$ for any $\eta$ which is zero at the boundary.
    \item $\frac{\delta^2 U[\phi^*]}{\delta \eta^2} > Dc||\eta||^2$ for some real number $c>0$ for any $\eta$ zero at the boundary, but not identically zero.
\end{itemize}
By the standard theory of variational calculus, $\phi^*$ is the unique global minimizer of \qref{nonlinear-energy} \cite{gelfand2000calculus}.
This completes the proof.

\section*{Acknowledgements}
The authors gratefully acknowledge financial support from Cummins Inc. under Grant No. 20067719.

\bibliographystyle{unsrtnat}
\bibliography{references}

\begin{thebibliography}{72}
\providecommand{\natexlab}[1]{#1}
\providecommand{\url}[1]{\texttt{#1}}
\expandafter\ifx\csname urlstyle\endcsname\relax
  \providecommand{\doi}[1]{doi: #1}\else
  \providecommand{\doi}{doi: \begingroup \urlstyle{rm}\Url}\fi

\bibitem[En{\ss}lin et~al.(2009)En{\ss}lin, Frommert, and
  Kitaura]{ensslin2009information}
Torsten~A En{\ss}lin, Mona Frommert, and Francisco~S Kitaura.
\newblock Information field theory for cosmological perturbation reconstruction
  and nonlinear signal analysis.
\newblock \emph{Physical Review D}, 80\penalty0 (10):\penalty0 105005, 2009.

\bibitem[En{\ss}lin(2013)]{ensslin2013information}
Torsten En{\ss}lin.
\newblock Information field theory.
\newblock In \emph{AIP Conference Proceedings}, volume 1553, pages 184--191.
  American Institute of Physics, 2013.

\bibitem[En{\ss}lin(2022)]{ensslin2022information}
Torsten En{\ss}lin.
\newblock Information field theory and artificial intelligence.
\newblock \emph{Entropy}, 24\penalty0 (3):\penalty0 374, 2022.

\bibitem[Raissi et~al.(2017)Raissi, Perdikaris, and
  Karniadakis]{raissi2017physics}
Maziar Raissi, Paris Perdikaris, and George~Em Karniadakis.
\newblock Physics informed deep learning (part i): Data-driven solutions of
  nonlinear partial differential equations.
\newblock \emph{arXiv preprint arXiv:1711.10561}, 2017.

\bibitem[Yang et~al.(2021)Yang, Meng, and Karniadakis]{yang2021b}
Liu Yang, Xuhui Meng, and George~Em Karniadakis.
\newblock B-pinns: Bayesian physics-informed neural networks for forward and
  inverse pde problems with noisy data.
\newblock \emph{Journal of Computational Physics}, 425:\penalty0 109913, 2021.

\bibitem[Stiasny et~al.(2021)Stiasny, Chevalier, and
  Chatzivasileiadis]{stiasny2021learning}
Jochen Stiasny, Samuel Chevalier, and Spyros Chatzivasileiadis.
\newblock Learning without data: Physics-informed neural networks for fast
  time-domain simulation.
\newblock In \emph{2021 IEEE International Conference on Communications,
  Control, and Computing Technologies for Smart Grids (SmartGridComm)}, pages
  438--443. IEEE, 2021.

\bibitem[Psichogios and Ungar(1992)]{psichogios1992hybrid}
Dimitris~C Psichogios and Lyle~H Ungar.
\newblock A hybrid neural network-first principles approach to process
  modeling.
\newblock \emph{AIChE Journal}, 38\penalty0 (10):\penalty0 1499--1511, 1992.

\bibitem[Meade~Jr and Fernandez(1994)]{meade1994numerical}
Andrew~J Meade~Jr and Alvaro~A Fernandez.
\newblock The numerical solution of linear ordinary differential equations by
  feedforward neural networks.
\newblock \emph{Mathematical and Computer Modelling}, 19\penalty0
  (12):\penalty0 1--25, 1994.

\bibitem[Lagaris et~al.(1998)Lagaris, Likas, and
  Fotiadis]{lagaris1998artificial}
Isaac~E Lagaris, Aristidis Likas, and Dimitrios~I Fotiadis.
\newblock Artificial neural networks for solving ordinary and partial
  differential equations.
\newblock \emph{IEEE transactions on neural networks}, 9\penalty0 (5):\penalty0
  987--1000, 1998.

\bibitem[Raissi and Karniadakis(2018)]{raissi2018hidden}
Maziar Raissi and George~Em Karniadakis.
\newblock Hidden physics models: Machine learning of nonlinear partial
  differential equations.
\newblock \emph{Journal of Computational Physics}, 357:\penalty0 125--141,
  2018.

\bibitem[Sahli~Costabal et~al.(2020)Sahli~Costabal, Yang, Perdikaris, Hurtado,
  and Kuhl]{10.3389/fphy.2020.00042}
Francisco Sahli~Costabal, Yibo Yang, Paris Perdikaris, Daniel~E. Hurtado, and
  Ellen Kuhl.
\newblock Physics-informed neural networks for cardiac activation mapping.
\newblock \emph{Frontiers in Physics}, 8:\penalty0 42, 2020.
\newblock ISSN 2296-424X.
\newblock \doi{10.3389/fphy.2020.00042}.
\newblock URL
  \url{https://www.frontiersin.org/article/10.3389/fphy.2020.00042}.

\bibitem[Kissas et~al.(2020)Kissas, Yang, Hwuang, Witschey, Detre, and
  Perdikaris]{KISSAS2020112623}
Georgios Kissas, Yibo Yang, Eileen Hwuang, Walter~R. Witschey, John~A. Detre,
  and Paris Perdikaris.
\newblock Machine learning in cardiovascular flows modeling: Predicting
  arterial blood pressure from non-invasive 4d flow mri data using
  physics-informed neural networks.
\newblock \emph{Computer Methods in Applied Mechanics and Engineering},
  358:\penalty0 112623, 2020.
\newblock ISSN 0045-7825.
\newblock \doi{https://doi.org/10.1016/j.cma.2019.112623}.
\newblock URL
  \url{http://www.sciencedirect.com/science/article/pii/S0045782519305055}.

\bibitem[Zhang et~al.(2020)Zhang, Yin, and
  Karniadakis]{zhang2020physicsinformed}
Enrui Zhang, Minglang Yin, and George~Em Karniadakis.
\newblock Physics-informed neural networks for nonhomogeneous material
  identification in elasticity imaging, 2020.

\bibitem[Cai et~al.(2021)Cai, Wang, Wang, Perdikaris, and
  Karniadakis]{cai2021physics}
Shengze Cai, Zhicheng Wang, Sifan Wang, Paris Perdikaris, and George~Em
  Karniadakis.
\newblock Physics-informed neural networks for heat transfer problems.
\newblock \emph{Journal of Heat Transfer}, 143\penalty0 (6):\penalty0 060801,
  2021.

\bibitem[Beltr{\'a}n-Pulido et~al.(2022)Beltr{\'a}n-Pulido, Bilionis, and
  Aliprantis]{beltran2022physics}
Andr{\'e}s Beltr{\'a}n-Pulido, Ilias Bilionis, and Dionysios Aliprantis.
\newblock Physics-informed neural networks for solving parametric magnetostatic
  problems.
\newblock \emph{arXiv preprint arXiv:2202.04041}, 2022.

\bibitem[Tartakovsky et~al.(2018)Tartakovsky, Marrero, Tartakovsky, and
  Barajas-Solano]{tartakovsky2018learning}
Alexandre~M Tartakovsky, Carlos~Ortiz Marrero, D~Tartakovsky, and David
  Barajas-Solano.
\newblock Learning parameters and constitutive relationships with physics
  informed deep neural networks.
\newblock \emph{arXiv preprint arXiv:1808.03398}, 2018.

\bibitem[Iten et~al.(2020)Iten, Metger, Wilming, del Rio, and
  Renner]{PhysRevLett.124.010508}
Raban Iten, Tony Metger, Henrik Wilming, L\'{\i}dia del Rio, and Renato Renner.
\newblock Discovering physical concepts with neural networks.
\newblock \emph{Phys. Rev. Lett.}, 124:\penalty0 010508, Jan 2020.
\newblock \doi{10.1103/PhysRevLett.124.010508}.
\newblock URL \url{https://link.aps.org/doi/10.1103/PhysRevLett.124.010508}.

\bibitem[Lu et~al.(2019)Lu, Jin, and Karniadakis]{lu2019deeponet}
Lu~Lu, Pengzhan Jin, and George~Em Karniadakis.
\newblock Deeponet: Learning nonlinear operators for identifying differential
  equations based on the universal approximation theorem of operators.
\newblock \emph{arXiv preprint arXiv:1910.03193}, 2019.

\bibitem[Marelli and Sudret(2014)]{marelli2014uqlab}
Stefano Marelli and Bruno Sudret.
\newblock Uqlab: A framework for uncertainty quantification in matlab.
\newblock In \emph{Vulnerability, uncertainty, and risk: quantification,
  mitigation, and management}, pages 2554--2563. 2014.

\bibitem[Karumuri et~al.(2019)Karumuri, Tripathy, Bilionis, and
  Panchal]{karumuri2019simulator}
Sharmila Karumuri, Rohit Tripathy, Ilias Bilionis, and Jitesh Panchal.
\newblock Simulator-free solution of high-dimensional stochastic elliptic
  partial differential equations using deep neural networks.
\newblock \emph{arXiv preprint arXiv:1902.05200}, 2019.

\bibitem[Zhu et~al.(2019)Zhu, Zabaras, Koutsourelakis, and
  Perdikaris]{zhu2019physics}
Yinhao Zhu, Nicholas Zabaras, Phaedon-Stelios Koutsourelakis, and Paris
  Perdikaris.
\newblock Physics-constrained deep learning for high-dimensional surrogate
  modeling and uncertainty quantification without labeled data.
\newblock \emph{Journal of Computational Physics}, 394:\penalty0 56--81, 2019.

\bibitem[Graves(2011)]{graves2011practical}
Alex Graves.
\newblock Practical variational inference for neural networks.
\newblock \emph{Advances in neural information processing systems}, 24, 2011.

\bibitem[Blundell et~al.(2015)Blundell, Cornebise, Kavukcuoglu, and
  Wierstra]{blundell2015weight}
Charles Blundell, Julien Cornebise, Koray Kavukcuoglu, and Daan Wierstra.
\newblock Weight uncertainty in neural network.
\newblock In \emph{International Conference on Machine Learning}, pages
  1613--1622. PMLR, 2015.

\bibitem[Gal and Ghahramani(2016)]{gal2016dropout}
Yarin Gal and Zoubin Ghahramani.
\newblock Dropout as a bayesian approximation: Representing model uncertainty
  in deep learning.
\newblock In \emph{international conference on machine learning}, pages
  1050--1059. PMLR, 2016.

\bibitem[Yang et~al.(2020)Yang, Zhang, and Karniadakis]{yang2020physics}
Liu Yang, Dongkun Zhang, and George~Em Karniadakis.
\newblock Physics-informed generative adversarial networks for stochastic
  differential equations.
\newblock \emph{SIAM Journal on Scientific Computing}, 42\penalty0
  (1):\penalty0 A292--A317, 2020.

\bibitem[Zhang et~al.(2019)Zhang, Lu, Guo, and
  Karniadakis]{zhang2019quantifying}
Dongkun Zhang, Lu~Lu, Ling Guo, and George~Em Karniadakis.
\newblock Quantifying total uncertainty in physics-informed neural networks for
  solving forward and inverse stochastic problems.
\newblock \emph{Journal of Computational Physics}, 397:\penalty0 108850, 2019.

\bibitem[MacKay(1992)]{mackay1992practical}
David~JC MacKay.
\newblock A practical bayesian framework for backpropagation networks.
\newblock \emph{Neural computation}, 4\penalty0 (3):\penalty0 448--472, 1992.

\bibitem[Neal(2012)]{neal2012bayesian}
Radford~M Neal.
\newblock \emph{Bayesian learning for neural networks}, volume 118.
\newblock Springer Science \& Business Media, 2012.

\bibitem[Neal et~al.(2011)]{neal2011mcmc}
Radford~M Neal et~al.
\newblock Mcmc using hamiltonian dynamics.
\newblock \emph{Handbook of markov chain monte carlo}, 2\penalty0
  (11):\penalty0 2, 2011.

\bibitem[Blei et~al.(2017)Blei, Kucukelbir, and McAuliffe]{blei2017variational}
David~M Blei, Alp Kucukelbir, and Jon~D McAuliffe.
\newblock Variational inference: A review for statisticians.
\newblock \emph{Journal of the American statistical Association}, 112\penalty0
  (518):\penalty0 859--877, 2017.

\bibitem[Bilionis(2016)]{bilionis2016}
Ilias Bilionis.
\newblock Probabilistic solvers for partial differential equations, 2016.
\newblock URL \url{https://arxiv.org/abs/1607.03526}.

\bibitem[Frank and En{\ss}lin(2020)]{frank2020probabilistic}
Philipp Frank and Torsten~A En{\ss}lin.
\newblock Probabilistic simulation of partial differential equations.
\newblock \emph{arXiv preprint arXiv:2010.06583}, 2020.

\bibitem[Chen et~al.(2021)Chen, Hosseini, Owhadi, and Stuart]{chen2021solving}
Yifan Chen, Bamdad Hosseini, Houman Owhadi, and Andrew~M Stuart.
\newblock Solving and learning nonlinear pdes with gaussian processes.
\newblock \emph{Journal of Computational Physics}, 447:\penalty0 110668, 2021.

\bibitem[Meng et~al.(2021)Meng, Yang, Mao, Ferrandis, and
  Karniadakis]{meng2021learning}
Xuhui Meng, Liu Yang, Zhiping Mao, Jose del~Aguila Ferrandis, and George~Em
  Karniadakis.
\newblock Learning functional priors and posteriors from data and physics.
\newblock \emph{arXiv preprint arXiv:2106.05863}, 2021.

\bibitem[Cotter et~al.(2010)Cotter, Dashti, and
  Stuart]{cotter2010approximation}
Simon~L Cotter, Masoumeh Dashti, and Andrew~M Stuart.
\newblock Approximation of bayesian inverse problems for pdes.
\newblock \emph{SIAM journal on numerical analysis}, 48\penalty0 (1):\penalty0
  322--345, 2010.

\bibitem[Dashti et~al.(2013)Dashti, Law, Stuart, and Voss]{dashti2013map}
Masoumeh Dashti, Kody~JH Law, Andrew~M Stuart, and Jochen Voss.
\newblock Map estimators and their consistency in bayesian nonparametric
  inverse problems.
\newblock \emph{Inverse Problems}, 29\penalty0 (9):\penalty0 095017, 2013.

\bibitem[Kullback(1997)]{kullback1997information}
Solomon Kullback.
\newblock \emph{Information theory and statistics}.
\newblock Courier Corporation, 1997.

\bibitem[Frewer et~al.(2016)Frewer, Khujadze, and Foysi]{frewer2016note}
Michael Frewer, George Khujadze, and Holger Foysi.
\newblock A note on the notion" statistical symmetry".
\newblock \emph{arXiv preprint arXiv:1602.08039}, 2016.

\bibitem[Parisi and Shankar(1988)]{parisi1988statistical}
Giorgio Parisi and Ramamurti Shankar.
\newblock Statistical field theory.
\newblock 1988.

\bibitem[Cartier and DeWitt-Morette(2006)]{cartier2006functional}
Pierre Cartier and C{\'e}cile DeWitt-Morette.
\newblock \emph{Functional integration: action and symmetries}.
\newblock Cambridge University Press, 2006.

\bibitem[Lancaster and Blundell(2014)]{lancaster2014quantum}
Tom Lancaster and Stephen~J Blundell.
\newblock \emph{Quantum field theory for the gifted amateur}.
\newblock OUP Oxford, 2014.

\bibitem[Treves(2016)]{treves2016topological}
Fran{\c{c}}ois Treves.
\newblock \emph{Topological Vector Spaces, Distributions and Kernels: Pure and
  Applied Mathematics, Vol. 25}, volume~25.
\newblock Elsevier, 2016.

\bibitem[Keener(2018)]{keener2018principles}
James~P Keener.
\newblock \emph{Principles of applied mathematics: transformation and
  approximation}.
\newblock CRC Press, 2018.

\bibitem[Albeverio et~al.(1976)Albeverio, H{\"o}egh-Krohn, and
  Mazzucchi]{albeverio1976mathematical}
Sergio Albeverio, Raphael H{\"o}egh-Krohn, and Sonia Mazzucchi.
\newblock \emph{Mathematical theory of Feynman path integrals}, volume 523.
\newblock Springer, 1976.

\bibitem[Bardeen et~al.(1986)Bardeen, Bond, Kaiser, and
  Szalay]{bardeen1986statistics}
James~M Bardeen, JR~Bond, Nick Kaiser, and AS~Szalay.
\newblock The statistics of peaks of gaussian random fields.
\newblock \emph{The Astrophysical Journal}, 304:\penalty0 15--61, 1986.

\bibitem[Kreyszig(1991)]{kreyszig1991introductory}
Erwin Kreyszig.
\newblock \emph{Introductory functional analysis with applications}, volume~17.
\newblock John Wiley \& Sons, 1991.

\bibitem[Deng(2011)]{deng2011generalization}
Chun~Yuan Deng.
\newblock A generalization of the sherman--morrison--woodbury formula.
\newblock \emph{Applied Mathematics Letters}, 24\penalty0 (9):\penalty0
  1561--1564, 2011.

\bibitem[Schulz et~al.(2018)Schulz, Speekenbrink, and
  Krause]{schulz2018tutorial}
Eric Schulz, Maarten Speekenbrink, and Andreas Krause.
\newblock A tutorial on gaussian process regression: Modelling, exploring, and
  exploiting functions.
\newblock \emph{Journal of Mathematical Psychology}, 85:\penalty0 1--16, 2018.

\bibitem[Welling and Teh(2011)]{welling2011bayesian}
Max Welling and Yee~W Teh.
\newblock Bayesian learning via stochastic gradient langevin dynamics.
\newblock In \emph{Proceedings of the 28th international conference on machine
  learning (ICML-11)}, pages 681--688, 2011.

\bibitem[Naki{\'c} and Veseli{\'c}(2020)]{nakic2020perturbation}
Ivica Naki{\'c} and Kre{\v{s}}imir Veseli{\'c}.
\newblock Perturbation of eigenvalues of the klein--gordon operators.
\newblock \emph{Revista Matem{\'a}tica Complutense}, 33\penalty0 (2):\penalty0
  557--581, 2020.

\bibitem[Tong(2017)]{tong2017statistical}
David Tong.
\newblock Statistical field theory.
\newblock \emph{Lecture Notes}, 2017.

\bibitem[Knollm{\"u}ller and En{\ss}lin(2019)]{knollmuller2019metric}
Jakob Knollm{\"u}ller and Torsten~A En{\ss}lin.
\newblock Metric gaussian variational inference.
\newblock \emph{arXiv preprint arXiv:1901.11033}, 2019.

\bibitem[Selig et~al.(2013)Selig, Bell, Junklewitz, Oppermann, Reinecke,
  Greiner, Pachajoa, and En{\ss}lin]{selig2013nifty}
Marco Selig, Michael~R Bell, Henrik Junklewitz, Niels Oppermann, Martin
  Reinecke, Maksim Greiner, Carlos Pachajoa, and Torsten~A En{\ss}lin.
\newblock Nifty--numerical information field theory-a versatile python library
  for signal inference.
\newblock \emph{Astronomy \& Astrophysics}, 554:\penalty0 A26, 2013.

\bibitem[Scarselli and Tsoi(1998)]{scarselli1998universal}
Franco Scarselli and Ah~Chung Tsoi.
\newblock Universal approximation using feedforward neural networks: A survey
  of some existing methods, and some new results.
\newblock \emph{Neural networks}, 11\penalty0 (1):\penalty0 15--37, 1998.

\bibitem[Robbins and Monro(1951)]{robbins1951stochastic}
Herbert Robbins and Sutton Monro.
\newblock A stochastic approximation method.
\newblock \emph{The annals of mathematical statistics}, pages 400--407, 1951.

\bibitem[Betancourt(2017)]{betancourt2017conceptual}
Michael Betancourt.
\newblock A conceptual introduction to hamiltonian monte carlo.
\newblock \emph{arXiv preprint arXiv:1701.02434}, 2017.

\bibitem[Dang et~al.(2019)Dang, Quiroz, Kohn, Minh-Ngoc, and
  Villani]{dang2019hamiltonian}
Khue-Dung Dang, Matias Quiroz, Robert Kohn, Tran Minh-Ngoc, and Mattias
  Villani.
\newblock Hamiltonian monte carlo with energy conserving subsampling.
\newblock \emph{Journal of machine learning research}, 20, 2019.

\bibitem[Chen et~al.(2014)Chen, Fox, and Guestrin]{chen2014stochastic}
Tianqi Chen, Emily Fox, and Carlos Guestrin.
\newblock Stochastic gradient hamiltonian monte carlo.
\newblock In \emph{International conference on machine learning}, pages
  1683--1691. PMLR, 2014.

\bibitem[Kingma and Ba(2014)]{kingma2014adam}
Diederik~P Kingma and Jimmy Ba.
\newblock Adam: A method for stochastic optimization.
\newblock \emph{arXiv preprint arXiv:1412.6980}, 2014.

\bibitem[Gelman et~al.(2013)Gelman, Carlin, Stern, Dunson, Vehtari, and
  Rubin]{gelman2013bayesian}
A.~Gelman, J.B. Carlin, H.S. Stern, D.B. Dunson, A.~Vehtari, and D.B. Rubin.
\newblock \emph{Bayesian Data Analysis, Third Edition}.
\newblock Chapman \& Hall/CRC Texts in Statistical Science. Taylor \& Francis,
  2013.
\newblock ISBN 9781439840955.
\newblock URL \url{https://books.google.com/books?id=ZXL6AQAAQBAJ}.

\bibitem[Phan et~al.(2019)Phan, Pradhan, and Jankowiak]{phan2019composable}
Du~Phan, Neeraj Pradhan, and Martin Jankowiak.
\newblock Composable effects for flexible and accelerated probabilistic
  programming in numpyro.
\newblock \emph{arXiv preprint arXiv:1912.11554}, 2019.

\bibitem[Courant(2005)]{courant2005dirichlet}
Richard Courant.
\newblock \emph{Dirichlet's principle, conformal mapping, and minimal
  surfaces}.
\newblock Courier Corporation, 2005.

\bibitem[Sargsyan et~al.(2019)Sargsyan, Huan, and Najm]{sargsyan2019embedded}
Khachik Sargsyan, Xun Huan, and Habib~N Najm.
\newblock Embedded model error representation for bayesian model calibration.
\newblock \emph{International Journal for Uncertainty Quantification},
  9\penalty0 (4), 2019.

\bibitem[Bilionis and Zabaras(2016)]{bilionis2016bayesian}
Ilias Bilionis and Nicholas Zabaras.
\newblock Bayesian uncertainty propagation using gaussian processes.
\newblock \emph{Handbook of uncertainty quantification}, pages 1--45, 2016.

\bibitem[Pacard(2009)]{pacard2009geometric}
Frank Pacard.
\newblock Geometric aspects of the allen--cahn equation.
\newblock \emph{Matematica Contempor{\^a}nea}, 37:\penalty0 91--122, 2009.

\bibitem[Davis and Sampson(1986)]{davis1986statistics}
John~C Davis and Robert~J Sampson.
\newblock \emph{Statistics and data analysis in geology}, volume 646.
\newblock Wiley New York, 1986.

\bibitem[Hoffman et~al.(2014)Hoffman, Gelman, et~al.]{hoffman2014no}
Matthew~D Hoffman, Andrew Gelman, et~al.
\newblock The no-u-turn sampler: adaptively setting path lengths in hamiltonian
  monte carlo.
\newblock \emph{J. Mach. Learn. Res.}, 15\penalty0 (1):\penalty0 1593--1623,
  2014.

\bibitem[Reynolds(2009)]{reynolds2009gaussian}
Douglas~A Reynolds.
\newblock Gaussian mixture models.
\newblock \emph{Encyclopedia of biometrics}, 741\penalty0 (659-663), 2009.

\bibitem[Feynman et~al.(2010)Feynman, Hibbs, and Styer]{feynman2010quantum}
Richard~P Feynman, Albert~R Hibbs, and Daniel~F Styer.
\newblock \emph{Quantum mechanics and path integrals}.
\newblock Courier Corporation, 2010.

\bibitem[Pandey et~al.(2022)Pandey, Singh, and Gardoni]{pandey2022review}
Aditya Pandey, Ashmeet Singh, and Paolo Gardoni.
\newblock A review of information field theory for bayesian inference of random
  fields.
\newblock \emph{Structural Safety}, 99:\penalty0 102225, 2022.

\bibitem[Poincar{\'e}(1890)]{poincare1890equations}
Henri Poincar{\'e}.
\newblock Sur les {\'e}quations aux d{\'e}riv{\'e}es partielles de la physique
  math{\'e}matique.
\newblock \emph{American Journal of Mathematics}, pages 211--294, 1890.

\bibitem[Gelfand et~al.(2000)Gelfand, Silverman, et~al.]{gelfand2000calculus}
Izrail~Moiseevitch Gelfand, Richard~A Silverman, et~al.
\newblock \emph{Calculus of variations}.
\newblock Courier Corporation, 2000.

\end{thebibliography}

\end{document}